\documentclass{article}
\usepackage[left=3cm, right=3cm, top=2cm]{geometry}
\usepackage{graphicx}
\usepackage{array}
\usepackage{nicefrac}
\usepackage{amsfonts}
\usepackage{mathtools}
\usepackage{hhline}
\usepackage{amsmath}
\usepackage{amsbsy}
\usepackage{accents}
\usepackage{textcomp}
\newcommand{\dbtilde}[1]{\accentset{\approx}{#1}}
\usepackage{moresize}
\newcolumntype{?}{!{\vrule width 1pt}}

\newcolumntype{P}[1]{>{\centering\arraybackslash}p{#1}}
\usepackage[table,xcdraw]{xcolor}

\usepackage{lineno}

\usepackage{multirow}
\usepackage{lipsum}                     
\usepackage{xargs}
\usepackage[colorinlistoftodos,prependcaption,textsize=tiny]{todonotes}
\newcommandx{\question}[2][1=]{\todo[linecolor=red,backgroundcolor=red!25,bordercolor=red,#1]{#2}}
\newcommandx{\comment}[2][1=]{\todo[linecolor=blue,backgroundcolor=blue!25,bordercolor=blue,#1]{#2}}
\newcommandx{\help}[2][1=]{\todo[linecolor=OliveGreen,backgroundcolor=OliveGreen!25,bordercolor=OliveGreen,#1]{#2}}
\newcommandx{\improvement}[2][1=]{\todo[linecolor=Plum,backgroundcolor=Plum!25,bordercolor=Plum,#1]{#2}}
\newcommandx{\thiswillnotshow}[2][1=]{\todo[disable,#1]{#2}}
\definecolor{dr}{rgb}{0.7, 0, 0.0}

\definecolor{aka}{rgb}{0.8, 0.0, 0.4}
\newcommand{\aka}[1] {{\color{aka} #1}}

\definecolor{navyblue}{rgb}{0.6, 0.0, 0.8}
\newcommand{\navyblue}[1] {{\color{navyblue} #1}}

\title{Fast and automated biomarker detection in breath samples\\ with machine learning
}

\author{Angelika Skarysz$^1{}^*$, Dahlia Salman$^2$, Michael Eddleston$^3$,\\ Martin Sykora$^4$, Eug\'{e}nie Hunsicker$^5$, William H Nailon$^6$, Kareen Darnley$^7$,\\ Duncan B McLaren$^6$, C L Paul Thomas$^2$ \& Andrea Soltoggio$^1{}^*$}
\date{
\small
$^1$Computer Science Department, School of Science, Loughborough University, Loughborough UK. \\
$^2$Centre for Analytical Science, School of Science, Loughborough University, Loughborough UK. \\
$^3$Pharmacology, Toxicology \& Therapeutics Unit, University of Edinburgh, Edinburgh, UK. \\
$^4$Centre for Information Management, School of Business and Economics, Loughborough University, UK. \\
$^5$Mathematical Sciences Department, School of Science, Loughborough University, Loughborough UK. \\
$^6$Edinburgh Cancer Centre, NHS Lothian, Edinburgh, UK. \\
$^7$Clinical Research Facility, Western General Hospital, NHS Lothian, Edinburgh, UK.\\
$^\ast$Correspondence to: A.Skarysz@lboro.ac.uk, A.Soltoggio@lboro.ac.uk.
}

\begin{document}

\maketitle


\begin{abstract}
Volatile organic compounds (VOCs) in human breath can reveal a large spectrum of health conditions and can be used for fast, accurate and non-invasive diagnostics. Gas chromatography-mass spectrometry (GC-MS) is used to measure VOCs, but its application is limited by expert-driven data analysis that is time-consuming, subjective and may introduce errors. 
We propose a system to perform GC-MS data analysis that exploits deep learning pattern recognition ability to learn and automatically detect VOCs directly from raw data, thus bypassing expert-led processing. 
The new proposed approach showed to outperform the expert-led analysis by detecting a significantly higher number of VOCs in just a fraction of time while maintaining high specificity. These results suggest that the proposed method can help the large-scale deployment of breath-based diagnosis by reducing time and cost, and increasing accuracy and consistency. \\\\
\end{abstract}

\section*{Introduction} 
A typical human breath sample is thought to contain thousands of volatile organic compounds (VOCs), which are the products of metabolic, catabolic and exogenous exposure processes occurring continuously in the human body \cite{Smolinska2014}. This makes breath a particularly interesting medium for metabolomics, which describes the individuals' specific phenotype and health status by measuring the metabolites present in the biological sample and changes in their expressions \cite{Smolinska2014, Hollywood2006}. Breathomics \cite{Rattray2014} (i.e.,\ breath metabolomics) can bring insight into all the metabolic processes in the body and thus provide comprehensive information about the organism's condition, additionally enabling non-invasive and rapid sample acquisition. Breath analysis has the potential to expand the range of diagnosis platforms for fast and accurate detection of a disease at an early stage, or for metabolic phenotyping, and so contributing to the development of precision medicine and treatment optimisation \cite{Rattray2014}. Due to such benefits, breathomics is currently an extensively researched area. Over the past few years, studies have applied breathomics for biomarker discovery and presented the relationships among the changes in VOC patterns and different types of diseases, including chronic obstructive pulmonary disease \cite{VanBerkel2010}, diabetes \cite{Li2017}, as well as breast \cite{Fuchs2010}, colorectal \cite{Altomare2012} and lung cancer \cite{Phillips2010}.

\begin{figure*}
\centering
\begin{tikzpicture}
   \node[anchor=south west,inner sep=0] (image) at (0,0) {
    \centerline{\includegraphics[width=0.72\linewidth]{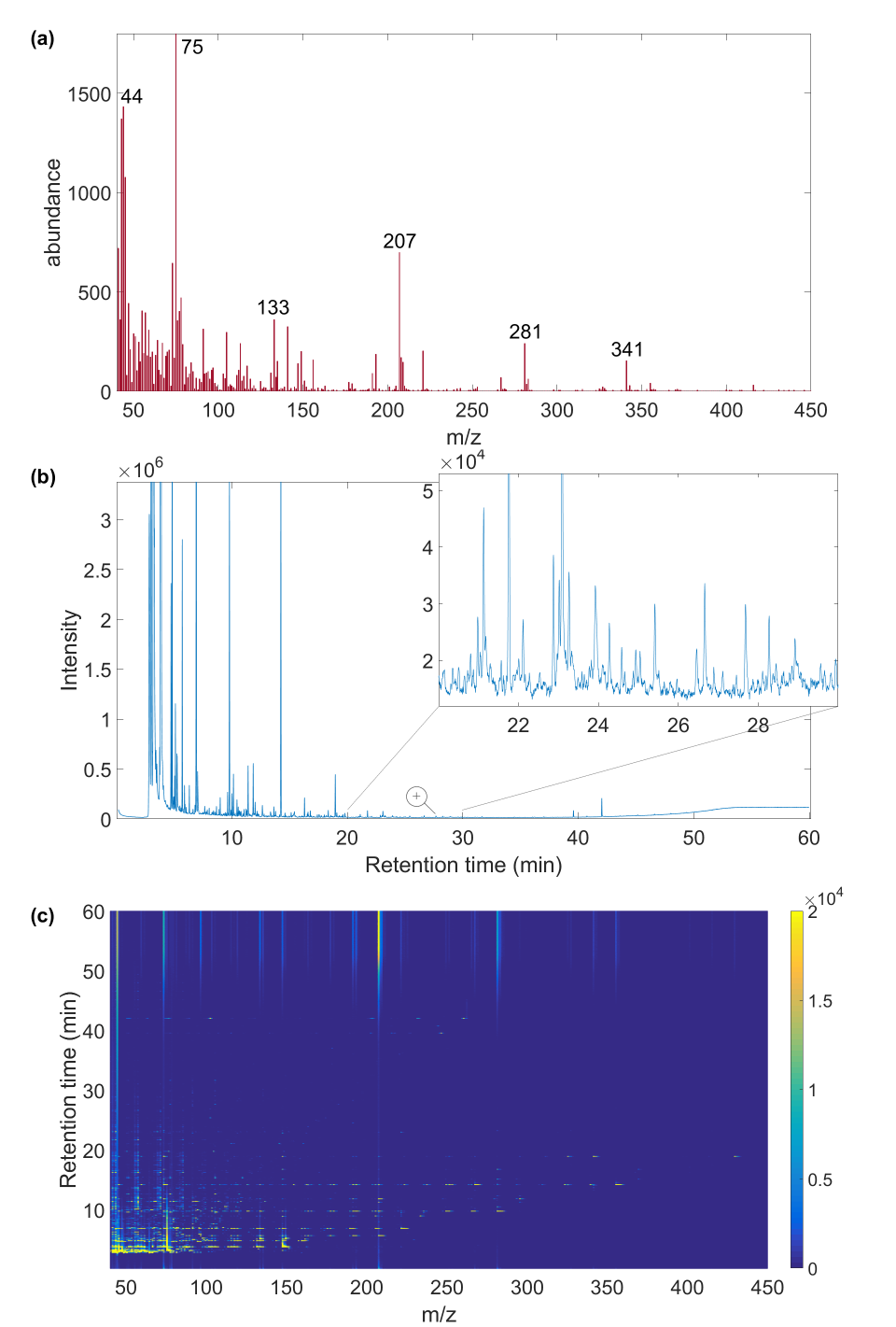}}};
    \begin{scope}[x={(image.south east)},y={(image.north west)}]
        \draw [-, thin, black] (0.344, 0.415) -- (0.49, 0.705);
    \end{scope}
\end{tikzpicture}
\caption{GC-MS breath data. (a) A full mass spectrum corresponds to each retention time (\emph{RT}) point on the chromatogram. (b) The total ion current (TIC) chromatogram plots along the \emph{RT} dimension the cumulative abundance of ions (intensity). Each peak generally represents one specific VOC,
although superposition of peaks also occurs \cite{Smolinska2014}. (c) GC-MS abundance matrix presented as a heat map, with the x-axis being the mass-to-charge ratio (\emph{m/z}) and the y-axis the retention time.
}
\label{fig:breathogram}
\end{figure*}

Accurate detection of the VOCs present in breath, in particular biomarkers related to a specific metabolic change caused by a disease, is essential to obtain a reliable diagnosis. Gas chromatography-mass spectrometry (GC-MS) is a well-known analytical technology that, due to its low limits of detection, orthogonal data structure and molecular structural characteristics, is the gold standard for VOC measurement in breath samples \cite{Watson2008}. Each VOC in the processed sample elutes from the GC analytical column with a retention time (\emph{RT}) related to its chemical and physical properties, but also dependent on the GC column features. That results in a separation of the compounds contained in the mixture of the sample. Subsequently, each eluted VOC is characterised in MS by a mass spectrum (mass-to-charge ratio, \emph{m/z}) of its ion fragmentation. As different VOCs produce different ion fragmentation patterns, an ion pattern enables the identification of the VOC. GC-MS produces a two-dimensional data matrix, known as abundance matrix \cite{stein1999integrated}, such as the one shown in Fig.\ \ref{fig:breathogram}c. Each column of the abundance matrix represents one ion channel \emph{m/z}, whereas each row contains a mass spectrum delivered at a specific \emph{RT}. GC-MS data is also often visualised compactly as total ion current (TIC) chromatogram
(Fig.\ \ref{fig:breathogram}b). For more details on GC-MS data, we refer readers to \cite{Hubschmann2015}. 

Deriving a comprehensive and reliable list of VOCs detected in GC-MS breath data is a difficult task. The VOC separation in GC may not fully occur, resulting in an overlapping of co-eluted compounds (i.e.,\ peaks on a chromatogram, Fig.\ \ref{fig:breathogram}b),
or the ion patterns produced in MS by some VOCs may be similar and thus difficult to distinguish. Moreover, a GC column degrades over time changing instrumentation features and thus causes \emph{RT} shifts, which significantly reduce the reliability of this parameter in VOC identification and requires its conversion into  a system-independent constant (i.e.,\ retention index) \cite{KovatsRI}. Additionally, the data produced by GC-MS are noisy and high dimensional: one single sample may contain over 9 million variables (in our study over 22500 \emph{RT} points by 411 \emph{m/z} channels). 

\begin{figure*}[htbp]
\centerline{\includegraphics[width=1\linewidth]{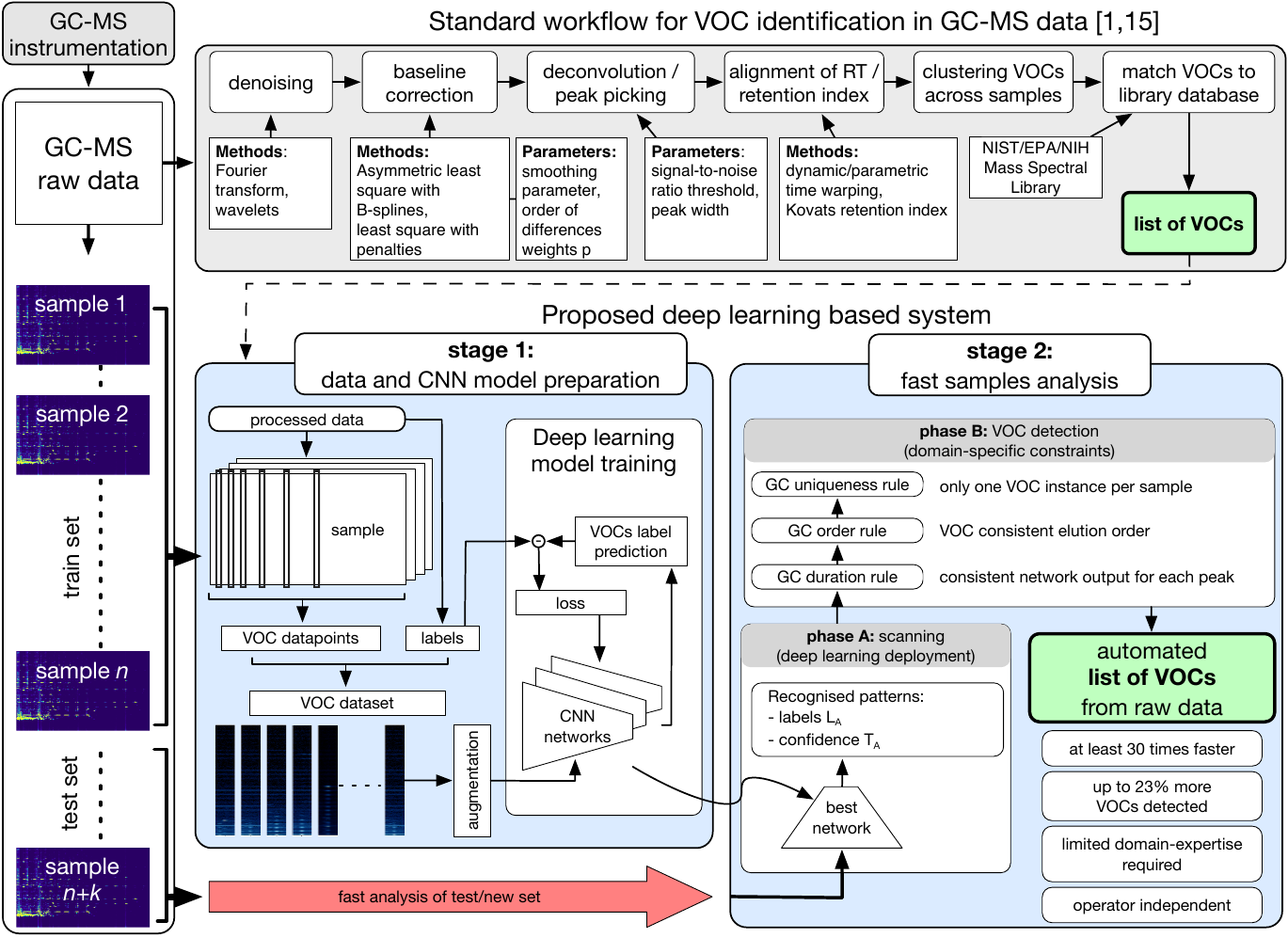}}
\caption{Graphical representation of the current GC-MS state-of-the-art analysis process (grey area) and the novel CNN-based method, stage 1 and stage 2. Both methods require GC-MS data storage (left column) and provide a list of VOCs as output.}
\label{fig:scheme}
\end{figure*}

For such complexity, the current breathomics workflow for GC-MS metabolic phenotyping employs various preprocessing steps, at the core of which is a spectral deconvolution, i.e.,\ an extraction of the overlapping co-eluted VOCs along with their mass spectra prints \cite{Colby1992}. 
The GC-MS breath data processing (Fig.\ \ref{fig:scheme}, top) includes baseline correction, spectral deconvolution, peak detection and feature alignment. These steps collectively enable the clustering and identification of the VOCs \cite{Alkhalifah2019} for further multivariate statistical analysis \cite{Smolinska2014, Ren2015}. These processes are subject to high variability. Usually, 350 to 500 VOCs are detected in the sample. The dynamic range of the variables may span $10^4$ to $10^5$ 
and some of the spectra for the lower abundance VOCs may be incomplete with minor ion fragments below the limits of detection. Variations in exogenous factors mean that different deconvolution approaches may need to be invoked from sample-to-sample. 
Current state-of-the-art deconvolution-based breathomics methods require analytical expertise and skilled analyst judgement to choose the techniques and parameters settings for processing the data from every sample. In particular, the  optimisation of deconvolution needs to strike the right balance between the inclusion of all the relevant VOCs and the omission of spectral noise and signal artefacts. 
The operator-subjective nature of the GC-MS data processing, combined with the data complexity, has the potential to introduce errors and variability to results on top of the observed biological variations from one subject to another, and thus the results may not be reproducible \cite{Coombes2007}. 
Additionally, expert-operated deconvolution is often labour-intensive time-consuming procedure \cite{Likic2009}; the processing time of a single breath sample is estimated by experts as 60 to 120 minutes.

The limitations outlined above call for better algorithms for GC-MS data processing, now possible by exploiting recent advances in machine learning and deep learning. A number of machine learning and deep learning applications in biomedical studies have been reported in the literature \cite{PaulSajda2006,Mamoshina2016}. Several studies, such as \cite{VanBerkel2010}, \cite{Altomare2012}, \cite{Baranska2013}, successfully applied machine learning in the area of breathomics. These reported applications, however, do not process GC-MS breath data directly but analyse a list of selected VOCs, provided by expert-led preprocessing, to classify patients and control group.
Consequently, they allow for high-level GC-MS data classification by black-box like decisions without justifications, rather than detection of particular VOCs of interest. Thus, these methods provide limited information about individual's metabolomics condition. Moreover, such data processing (e.g.,\ \cite{VanBerkel2010}, \cite{Altomare2012}, \cite{Baranska2013}) involves treating an entire GC-MS sample as a single data point; in the case of clinical samples, obtaining large-scale datasets desired for machine learning applications may be a challenging and demanding task. On the other hand, to the best of the authors' knowledge, there is a lack of studies applying machine learning directly on raw GC-MS breath data to detect VOCs of interest.

This study introduces a new approach to VOC detection in GC-MS samples: we propose the application of convolutional neural networks (CNNs) \cite{LeCun1989} to learn to detect VOCs in breath sample automatically and directly from raw GC-MS data, thereby bypassing the current labour-intensive, time-consuming and operator-subjective data preprocessing steps. CNNs \cite{Rawat2017} are a popular type of deep learning algorithms that is particularly effective in image analysis (e.g.,\ \cite{LeCun2004}, \cite{Ciresan2012}, \cite{Krizhevsky2012}). CNNs can autonomously learn useful features directly from low-level data, e.g.,\ pixels \cite{Sermanet2013}, and construct high-level features without human intervention. CNNs can also exploit geometrical properties of the data and thus adapt well to image-based tasks \cite{Nielsen2015}. Presenting GC-MS data as abundance matrix enables one to see GC-MS data as an image (Fig.\ \ref{fig:breathogram}c). Therefore, we propose a new approach, exploiting recent advancement in CNN applications for image analysis and pattern recognition, to learn to recognise ion patterns directly from raw abundance matrix. Ion patterns derived from specific compounds, although noisy, present unique features that distinguish them. A recognised ion pattern is effectively a recognised VOC, which in turn could be a biomarker of a given physical condition \cite{Garcia2011}.

The promise of such an approach was first demonstrated in \cite{Skarysz2018}. In that study, CNNs were shown to have considerably better performance than support vector machines \cite{Cortes1995} and shallow neural networks \cite{Zhang2000}. However, the study reported a high number of false positives and targeted only 8 VOCs, thus leaving open the question of reliability in detection and scalability.

The CNN-based approach proposed here has two main stages (Fig.\ \ref{fig:scheme}, bottom); stage 1: data and CNN model preparation, 
and stage 2: raw GC-MS samples analysis.
In stage 1, expert knowledge is exploited to create a dataset of target VOCs and their corresponding ion patterns on the raw GC-MS data. A CNN architecture is then trained on such dataset to learn to recognise ion patterns specific to the target VOCs.
In stage 2, new raw breath samples are analysed to detect the targeted VOCs quickly and automatically. Firstly (in phase A, Fig.\ \ref{fig:scheme}), an entire GC-MS sample is scanned by the trained CNN network to provide a list of recognised patterns. Secondly (in phase B), domain-specific constraints are used to derive a final list of detected VOCs.

The robustness and scalability of the proposed CNN-based method were investigated on a dataset of 120 GC-MS samples and the set of 30 target VOCs. A wide range of target compounds generates various challenges: several target VOCs produce similar ion fragmentation patterns, some others have close \emph{RT} positions and thus overlap, which may provide an obstacle to their accurate discrimination.
In this study we tested different types of CNNs: VGG16 and VGG-like networks \cite{Simonyan2014}, residual neural networks \cite{He2016} and densely connected convolutional networks \cite{huang2017densely} with different configurations, to compare their performance and select the most efficient strategy for GC-MS data processing.

The novel CNN-based approach proposed here provides a comprehensive system for fast and automated detection of any set of VOCs in raw GC-MS data, outperforming current expert-led deconvolution-based methods.
The analysis of raw GC-MS breath data may reduce human-related errors 
and has the potential to detect compounds of very low abundances. Consequently, the proposed novel approach showed the ability to correctly detect VOCs missed by the current methods, while improving specificity and significantly reducing processing time.
The system may support experts to put much more accurate hypotheses on the VOCs related to the specific health conditions. Moreover, by the significant acceleration of the VOC detection process, the CNN-based method allows for much quicker hypotheses validation on new GC-MS samples. To the best of the authors’ knowledge, this is the first study that proposes a comprehensive system to reveal VOC ion patterns directly from raw GC-MS samples, with high sensitivity and specificity.

\section*{Results} 

\subsection*{GC-MS sample dataset}

Breath samples were obtained in a clinical trial from 25 participants with different types of cancer receiving radiotherapy treatment. Four breath samples (prior and 1, 3, and 6 hours post radiation) were collected from each participant along with one environmental sample.

Each clinical sample was processed with GC-MS and stored in a data file containing both metadata and the abundance matrix $A\in\mathbb{R}^{\emph{R}\times 411}$, where 411 is the number of measured \emph{m/z} channels and \emph{R} is retention time dimension of approximately 22500 points. Subsequently, all GC-MS data were processed with the current expert-led methods to identify VOCs contained in the sample and their \emph{RT} positions. This process was completed for 120 clinical samples (see Methods).

The GC-MS sample dataset was divided into training and testing sets in the proportion 82/38: the training set contains 65 breath samples and 17 environmental samples associated with 17 randomly chosen participants, the testing set contains 30 breath samples and 8 environmental samples associated with remaining 8 participants. Both raw and expert-processed GC-MS samples from the training set were used to generate a VOC dataset (described later) for the CNN model training in stage 1 (Fig.\ \ref{fig:scheme}). The raw GC-MS samples from the testing set were used as inputs for automated VOC detection in stage 2, whereas their expert-processed equivalents made a ground truth for the proposed CNN-based system evaluation.

\subsection*{Target VOCs}
A set of 30 target VOCs (Table \ref{tabComposition}) was designed to contain compounds commonly found in the breath, including alkanes, aldehydes, ketones, furans and siloxanes and sulfur-containing compounds \cite{Alkhalifah2019}. The confidence in the expert-led identification process for these VOCs varies according to potential confounding factors. In particular, confounding factors include low concentrations of a VOC in a sample, which results in a low signal-to-noise ratio in GC-MS data. Propionic acid is an example of a target VOC reporting relatively low concentration in the clinical samples from the dataset. Other confounding factors are mass spectra overlapping caused by the co-elution of VOCs from the GC column, and the similar mass spectra among VOCs. For example, Octane and Hexanal frequently overlap, additionally sharing three of the five top ions. What is more, octane produces highly similar mass spectrum profiles as another target VOC -- 2,4-dimethylheptane. These VOCs also elute at a relatively close \emph{RT} locations and thus may be difficult to differentiate with the current expert-led processing \cite{Alkhalifah2019}. As a consequence of the abovementioned challenges, the process of expert-led VOC identification cannot be guaranteed to be error-free, resulting in a possibly noisy-labelled VOC dataset (stage 1) and ground truth (stage 2). The Supplementary Table 1 provides, for each target VOC, the details of its ion pattern, mean \emph{RT} location and mean concentration in GC-MS sample dataset, and a number of VOC instances reported respectively in breath and environmental samples.

\begin{table*}[htbp]
\caption{List of the target VOCs with class labels in the elution order. Blue frames indicate pairs of overlapping compounds. Compounds of relatively low concentrations are marked with colours: \aka{Mean EIC-Area} $\leq 10^3 \leq$ \navyblue{Mean EIC-Area} $\leq Q_1$ (first quartile); see Supplementary Table 1.}
\begin{center}
\begin{minipage}[b]{1 \linewidth}\centering
\begin{tabular}{|c|p{6cm}?c|p{6cm}|}
    \hline
    Label & Target compound & Label & Target compound\\
    \hline
    0 & Negative class & \textcolor{aka}{16} & \textcolor{aka}{Heptanal}\\
    1 & Ethanol & 17 & Benzaldehyde\\
    2 & Dimethyl sulfide & 18 & Benzonitrile\\ 
    3 & 2-Methylfuran & \textcolor{aka}{19} & \textcolor{aka}{Octanal}\\
    4 & Trichloromethane-d & 20 & Limonene\\
    5 & Benzene & 21 & 2-Ethylhexanol\\
    \textcolor{aka}{6} & \textcolor{aka}{Propionic acid} & 22 & Nonanal\\
    7 & Dimethyl disulfide & 23 & Dodecane\\
    8 & Toluene-D8 & \textcolor{aka}{24} & \textcolor{aka}{Decanal}\\
    9 & Toluene & 25 & 2-Phenoxyethanol\\
    10 & 3-Methylthiophene & \textcolor{navyblue}{26} & \textcolor{navyblue}{Phthalic acid}\\
    \textcolor{navyblue}{11} & \textcolor{navyblue}{Octane} & 27 & Tetradecane\\ 
    \textcolor{navyblue}{12} & \textcolor{navyblue}{Hexanal} & 28 & 1,4-Diacetylbenzene\\ 
        13 & 2,4-Dimethylheptane & \textcolor{aka}{29} & \textcolor{aka}{3,3,6,6-Tetraphenyl-1,2,4,5-tetroxane}\\
    14 & 2,4-Dimethyl-1-heptene & 30 & 2,5-Diphenyl-1,4-benzoquinone\\ 
    15 & 3-Heptanone & & \hspace{0.3cm} \\\hline
\end{tabular}
\label{tabComposition}
\end{minipage}
\end{center}
\begin{tikzpicture}[overlay]
\begin{scope}[x={(image.south east)},y={(image.north west)}]
\draw [-, semithick, blue] (0.02, 0.382) -- (0.39, 0.382);
\draw [-, semithick, blue] (0.02, 0.329) -- (0.39, 0.329);
\draw [-, semithick, blue] (0.02, 0.382) -- (0.02, 0.329);
\draw [-, semithick, blue] (0.39, 0.382) -- (0.39, 0.329);

\draw [-, semithick, blue] (0.02, 0.255) -- (0.39, 0.255);
\draw [-, semithick, blue] (0.02, 0.202) -- (0.39, 0.202);
\draw [-, semithick, blue] (0.02, 0.255) -- (0.02, 0.202);
\draw [-, semithick, blue] (0.39, 0.255) -- (0.39, 0.202);

\draw [-, semithick, blue] (0.02, 0.179) -- (0.39, 0.179);
\draw [-, semithick, blue] (0.02, 0.126) -- (0.39, 0.126);
\draw [-, semithick, blue] (0.02, 0.179) -- (0.02, 0.126);
\draw [-, semithick, blue] (0.39, 0.179) -- (0.39, 0.126);

\draw [-, semithick, blue] (0.515, 0.355) -- (0.885, 0.355);
\draw [-, semithick, blue] (0.515, 0.302) -- (0.885, 0.302);
\draw [-, semithick, blue] (0.515, 0.302) -- (0.515, 0.355);
\draw [-, semithick, blue] (0.885, 0.302) -- (0.885, 0.355);
 \end{scope}
\end{tikzpicture}
\end{table*}

\subsection*{VOC dataset} \label{dataset}

The proposed CNN-based approach relies on the construction of a dataset of target VOC patterns, derived from raw GC-MS samples for the network training. Each VOC appears on the TIC chromatogram (Fig.\ \ref{fig:breathogram}b) as one peak (sometimes overlapped) over a small segment of \emph{RT} axis, corresponding to a specific range of retention times when the VOC was eluting from the GC column. A sub-matrix of abundance matrix $A$, encompassing such an \emph{RT} range, contains the ion pattern for that specific VOC. Thus, we defined a VOC data point as a matrix $\hat{s}\in\mathbb{R}^{\delta\times 411}$, where the retention time window size was selected as $\delta=80$ (see Methods).

\begin{figure*}
\centerline{\includegraphics[width=0.9\linewidth]{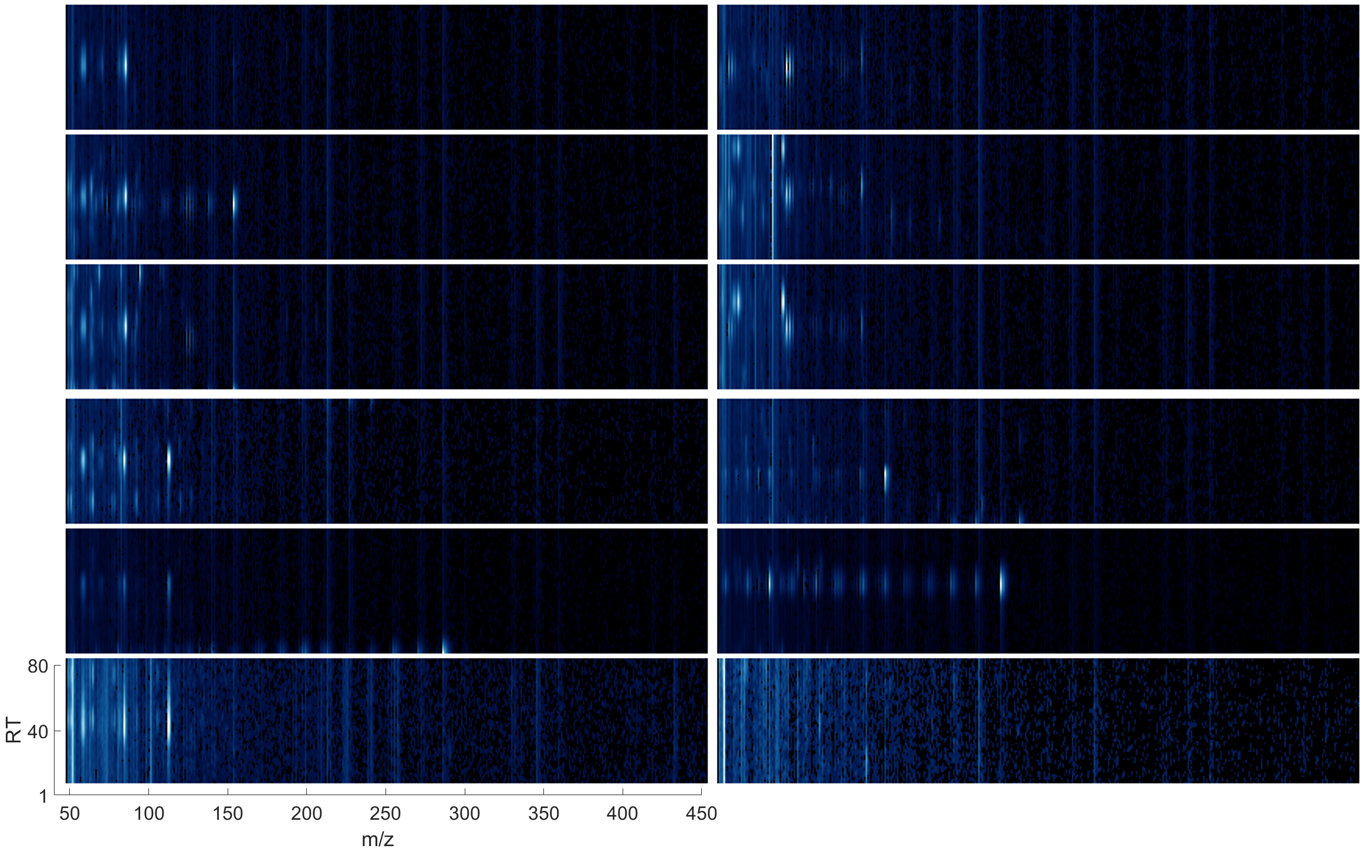}}
\caption{Examples of the VOC data points extracted from the raw GC-MS abundance matrices. Left, from top: three examples of Benzene, three examples of Benzaldehyde; right, from top: three examples of Trichloromethane-d and three example data points from the negative class. For better visualisation, the segments were mapped to RGB format.
}
\label{fig:segments}
\end{figure*}

Data points corresponding to all target VOC instances, previously identified by expert-led processing in the training GC-MS samples, were extracted to form the labelled VOC dataset for network training. In addition to data points representing target VOCs, a negative class of data points was created from randomly chosen sub-matrices $\hat{s}$ of $A$ that did not contain target VOCs. In total, from the 82 training GC-MS samples, 3,736 VOC data points were extracted with a minimum of 22 and a maximum of 82 data points in each class (Supplementary Table 1). Fig.\ \ref{fig:segments} shows examples of VOC data points extracted from raw GC-MS data.

\subsection*{Data augmentation and normalisation}

Data augmentation, i.e.,\ methods for enlargement of the dataset by insertion of unobserved data examples \cite{Dyk2001}, is known to benefit machine learning models where data points are scarce \cite{Dyk2001}. Often these new examples are constructed from the observed ones, by the introduction of some variations to data points, which do not change their underlying distribution (here, VOC ion patterns) and classes. 

Data augmentation was applied to the VOC dataset to increase the robustness of the training. 
We introduced two methods for VOC dataset augmentation: \emph{translation along RT} and \emph{intensity variation} (see Methods). Combined collectively, these two methods generated 100 variations for each VOC data point, resulting in the fully-augmented dataset of 373,600 VOC data points. To test the impact of the proposed augmentation methods and select the most efficient approach for network training, we compared the performance achieved on the fully-augmented dataset, partially-augmented dataset (after translation along \emph{RT} only) and original VOC dataset. The intensity values in each data point were normalised in the range \([0, 1]\).

\subsection*{Deep learning models}

We assessed the learning and inference capabilities of the following recently developed deep CNN architectures, which achieved state-of-the-art performance in several image classification challenges: VGG16 \cite{Simonyan2014}, residual CNNs \cite{He2016} and densely connected CNNs \cite{huang2017densely}. We also tested smaller VGG-like networks with 4, 6 and 8 layers.

Along the \emph{m/z} axis the values of a GC-MS abundance matrix are spatially only weakly correlated, as opposed to the data typically used in image-based tasks. Therefore, we tested the effectiveness of 1D filters (see Methods) and compared them with typically used 2D filters to investigate the suitability of such a network variation to capture the specific nature of correlation in GC-MS data. The details on tested CNN architectures, their implementations as well as parameters settings are given in Supplementary Tables 2.1-2.11.

\subsection*{CNN model training}

To evaluate the CNN models' ability to learn ion patterns from the VOC dataset, and to select the best-performing configurations of the networks, five-fold cross-validation (CV) was applied. CV is a common technique used for model assessment and hyperparameter selection \cite{Hastie2009}. Each split of the VOC dataset for CV was performed at the level of participant.

Table \ref{CVresults} presents the classification performance of the models in the CV stage. All tested models achieved high accuracy (i.e.,\ a proportion of data points for which the network label matched the ground truth), which confirms that ion patterns can be learned by CNNs from VOC data points derived directly from raw GC-MS data. The results also validate the hypothesis that 1D filters are more effective than standard 2D filters. Among the tested augmentation strategies, the best models' performances were obtained with the fully-augmented dataset. Shallower architectures, i.e.,\ VGG-like, performed no worse than other deeper networks. The VGG-like network with 8 layers and 1D filters (VGG-8-1D) reported the highest accuracy. 

The best performing hyperparameter configuration for each investigated CNN architecture type, i.e.,\ VGG-8-1D, DenseNet-40-1D and ResNet-34-1D, was used to create the system for automated VOC detection in raw GC-MS samples (Fig.\ \ref{fig:scheme}, stage 2). Before the deployment to stage 2, the models were retrained on the entire, fully-augmented VOC dataset.

\begin{table*}[htbp]
\caption{Five-fold cross-validation (CV) performance achieved by tested CNN models on the VOC dataset. (Model VGG16 with 1D filters has not been tested due to its extensive demand for memory resources.)}
\begin{center}
\begin{tabular}{|c|c|c|P{2.5cm}|P{2.5cm}|P{2.5cm}|}
	\cline{4-6}
    \multicolumn{3}{c|}{} & \multicolumn{3}{c|}{CV accuracy: mean ($\pm$ std) \%}\\\hline
    Model & Depth & Filters & Fully-aug. & Partially-aug. & Original data\\
    \hline
    VGG-like & 4 & 1D & 97.16 ($\pm$0.82) & 96.89 ($\pm$0.65) & 94.60 ($\pm$1.02) \\
    VGG-like & 4 & 2D & 97.13 ($\pm$0.92) & 96.48 ($\pm$0.89) & 93.89 ($\pm$1.06)\\
    VGG-like & 6 & 1D & 97.60 ($\pm$1.08) & 97.43 ($\pm$0.78) & 94.75 ($\pm$1.11)\\
    VGG-like & 6 & 2D & 97.37 ($\pm$0.95) & 96.69 ($\pm$0.79) & 93.73 ($\pm$1.12)\\
    \textbf{VGG-like} & \textbf{8} & \textbf{1D} & \textbf{98.06 ($\pm$0.38)} & 97.83 ($\pm$0.38) & 94.51 ($\pm$1.43)\\
    VGG-like & 8 & 2D & 97.34 ($\pm$0.94) & 96.89 ($\pm$1.03) & 88.11 ($\pm$1.62)\\
    VGG16 & 16 & 2D & 97.89 ($\pm$0.47) & 97.68 ($\pm$0.40) & 96.14 ($\pm$1.63)\\\hline
    DenseNet & 40 & 1D & 97.34 ($\pm$1.18) & 96.51 ($\pm$1.67) & 94.55 ($\pm$2.69)\\
    DenseNet & 40 & 2D & 96.73 ($\pm$0.82) & 95.92 ($\pm$1.98) & 69.98 ($\pm$1.96)\\\hline
    ResNet & 34 & 1D & 97.71 ($\pm$0.68) & 97.64 ($\pm$0.45) & 94.26 ($\pm$2.50)\\
    ResNet & 34 & 2D & 97.25 ($\pm$0.88) & 96.98 ($\pm$0.67) & 92.84 ($\pm$1.53)\\\hline
\end{tabular}
\label{CVresults}
\end{center}
\end{table*}

\subsection*{Automated samples analysis and VOC detection}

The trained networks were employed in stage 2 (Fig.\ \ref{fig:scheme}) to analyse the 38 raw GC-MS samples from the testing set. The proposed stage 2 analysis was composed of two phases: (A) scanning of each clinical sample with the network; (B) identification of VOC detections from the scan results.

\textbf{\textsl{Phase A -- scanning of a breath sample:}} 
The abundance matrix $A$ of a clinical sample was scanned along the \emph{RT} dimension. Precisely, the network was fed consecutive normalised sub-matrices $s_i \in \mathbb{R}^{\delta\times 411}$ of $A$, starting at retention point $i$ (for each $i$ in the \emph{RT} dimension for that sample).
Phase A classifies each $s_i$ into one of the 31 possible classes. 
Thus, scanning a single GC-MS sample required approximately 22500 queries of the network. The output of the process are two sequences: a sequence $L_A$ that contained approximately 22500 class labels, and a sequence $T_A$ that contained the classification confidence given by the network for each respective label (see Methods).

\textbf{\textsl{Phase B -- VOC detection:}} To obtain a list of target VOCs detected in each clinical GC-MS sample, the pair of phase A output vectors for that sample was analysed. The following general properties of the GC-MS samples are considered in the phase B analysis:

\begin{itemize}
\item[(i)] Along the \emph{RT} dimension, a VOC is measured multiple times consecutively for the duration of the VOC elution (typically about 6 seconds). 
\item[(ii)] The order of elution of VOCs from the GC column is usually constant across different samples, as indicated in Table \ref{tabComposition}.
\item[(iii)] Each VOC elutes from the GC column only once in a GC-MS process.
\end{itemize}

According to (i), a detection $d_j \in \mathcal{D}_A$ was defined as a consecutive sub-sequence of $L_A$ of length at least $\gamma$ with constant label values $j$ (\emph{duration rule}, see Methods). Subsequently, according to (ii), the system ignored detections that were far from the expected order (\emph{order rule}), giving a set $\tilde{\mathcal{D}}_A \subset \mathcal{D}_A$. Finally, according to (iii), if multiple detections of one VOC occurred in any sample, the system selected the one with the highest \emph{detection confidence}, derived from $T_A$, expressing the confidence of the model in such detection (\emph{uniqueness rule}); a set $\dbtilde{\mathcal{D}}_A \subset \tilde{\mathcal{D}}_A$.

\begin{figure*}
    \centering
    \centerline{\includegraphics[width=1.0\linewidth]{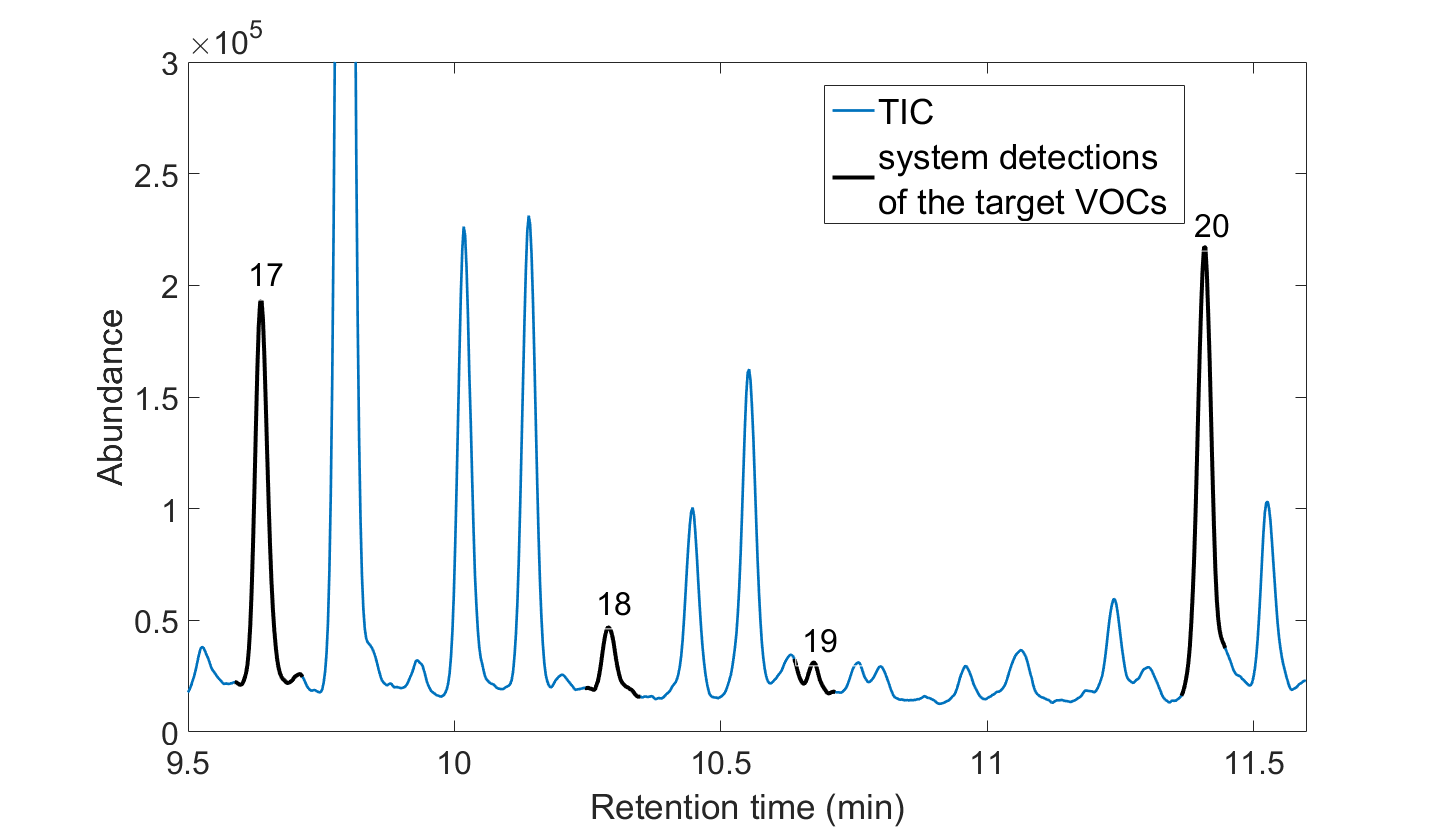}}
    \vspace{0.3cm}
    \begin{tabular}{|c|P{3.5cm}|P{3.5cm}|P{3.5cm}|}
    \hline
    VOC Label & Start detection $s\mathcal{RT}$ & End detection $e\mathcal{RT}$ & Confidence $\mathcal{T}$\\
    \hline
    17 & 9.588 & 9.712 & 1.0000\\
    18 & 10.243 & 10.359 & 1.0000\\
    19 & 10.638 & 10.715 & 0.9974\\ 
    20 & 11.364 & 11.447 & 1.0000\\\hline
    \end{tabular}
    \caption{Example output from clinical sample scanning for VOC detection (stage 2) with a CNN-based system, range between 9.5 and 11.5 minutes; sample Test-01-BS01, model VGG-8-1D (Supplementary Table 3.2.1). Top: VOCs detected by the system in this RT range, visualised on the TIC chromatogram. Bottom: Table of detected VOCs along with their RT positions and detection confidences.}
    \label{scan_figure}
\end{figure*}

As a result, for each abundance matrix $A$, stage 2 delivered a list of detected VOCs along with their location along the \emph{RT} axis and the detection confidence. A graphical representation of the output is provided in Fig.\ \ref{scan_figure}.

\subsection*{Evaluation of the automated sample analysis and VOC detection}

The performance of the system was assessed with two approaches: (1) including both the VOC localisation (along \emph{RT} dimension) and its classification; and additionally, to report the system specificity, (2) including only the VOC presence in a sample.

\textbf{\textsl{(1) -- VOC localisation and classification:}}
VOC detection involves both localisation of the compound in the GC-MS sample and its classification. Therefore, each VOC detection in a sample was identified as a true positive (TP) if there was matching in both label and \emph{RT} position with the ground truth for that sample, and as a false positive otherwise. Similarly, if a specific target VOC was not detected at the \emph{RT} position reported by the ground truth, it was identified as a false negative. Note that
a retention time window of any size in which the system and the expert both did not identify any target VOC could be considered as a true negative: therefore true negatives were not measured here.

\begin{figure*}
\centerline{\includegraphics[width=0.8\linewidth]{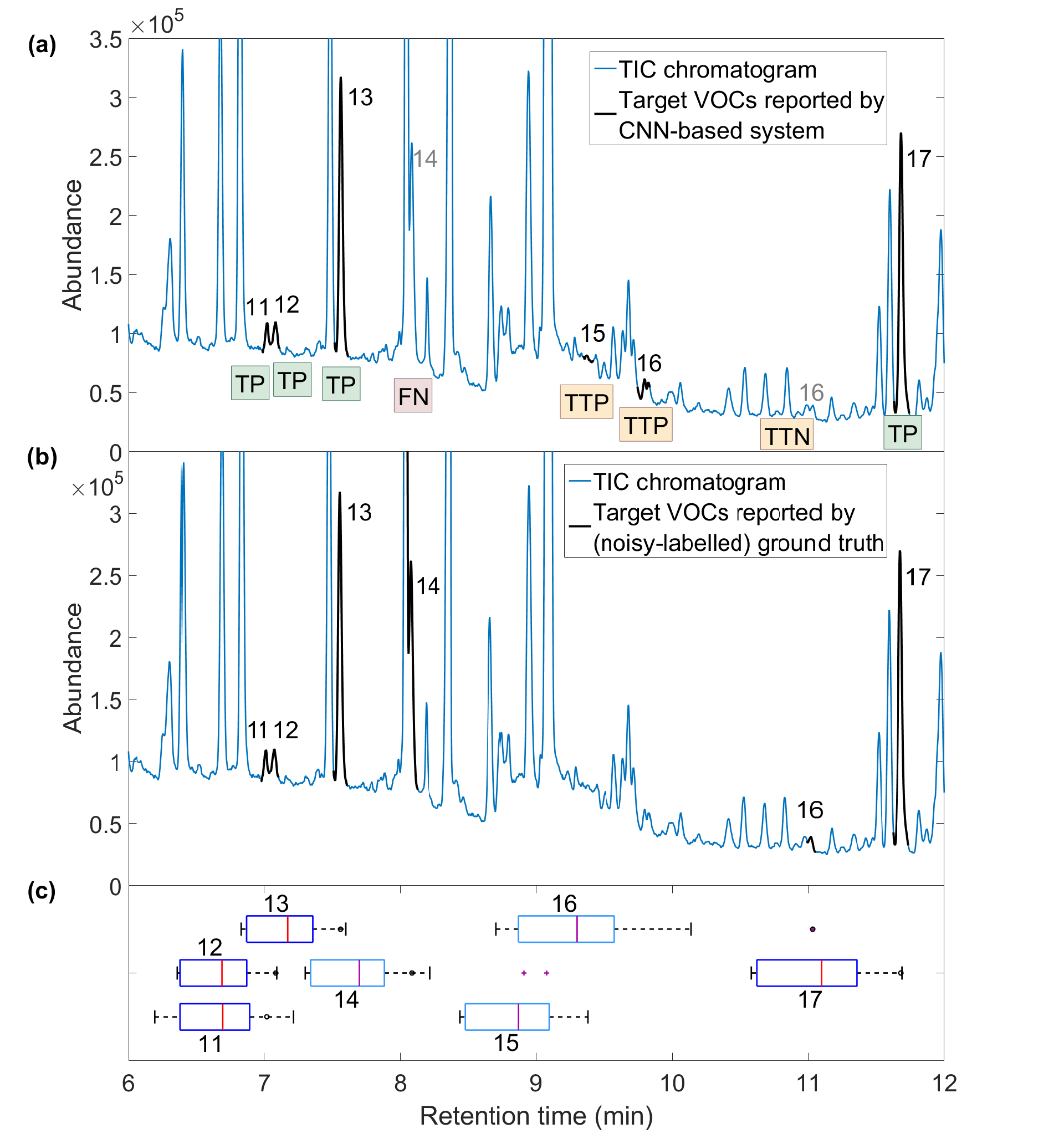}}
\caption{An evaluation of the results of a GC-MS sample scanning for VOC detection (stage 2) with the CNN-based system, range between 6 and 12 minutes; sample Test-04-BS01, model VGG-8-1D (Supplementary Table 3.2.16). (a) VOCs detected by the system presented on total ion current (TIC) chromatogram and marked accordingly to the outcome of the evaluation. VOCs 11, 12, 13 and 17 detected alike by the system and the ground truth, true positives. VOC 14 reported by the ground truth and not detected by the system, false negative. VOC 15 not reported by the ground truth, but detected by the system in compatible \emph{RT} range, tentative true positive. VOC 16 not detected by the system on the position reported by the ground truth, but detected on a different position within its RT range, tentative true negative and tentative true positive. (b) VOCs reported by the (noisy-labelled) ground truth presented on TIC chromatogram. (c) RT ranges specific for the VOCs, derived from the ground truth. Black dots indicate the exact RT positions of VOCs in the ground truth. Note that VOC 16 is reported by the ground truth at \emph{RT} position being an outlier for the \emph{RT} range of this compound.}
\label{fig:scan_figure2}
\end{figure*}

\textsl{System-derived ground truth corrections:}
To gain further insight on the results, we examined all the detections $\hat{d}_j \in \mathcal{D}_A$ (i.e.,\ before applying order and uniqueness rules) that were not reported in the ground truth or had mismatching \emph{RT} positions. 
The range of retention times in the ground truth was calculated for each VOC (see Methods). We observed that in most cases, the detection $\hat{d}_j$ was reported by the system in the expected (compatible) narrow \emph{RT} range for that VOC $j$.
The upper bound probability of a random false positive detection occurring within a precise and restricted \emph{RT} range was calculated as 4\% (see Methods). Accordingly, it is highly likely that such false positive is actually a true positive and reveals an error in the (noisy-labelled) ground truth, i.e.,\ a VOC occurrence missed in the expert-led analysis.
False positives with a compatible \emph{RT} value were named \emph{tentative true positives} (TTP), whereas false positives that are not TTP are called certain false positives (FP); see Fig.\ \ref{fig:scan_figure2}.

In some cases, the VOCs detected by the system and identified as TTP were, in fact, reported by the ground truth, but at different \emph{RT} positions (Fig.\ \ref{fig:scan_figure2}). Accepting the high chance of tentative true positives to be true positives and the constraints that a VOC appears only once in a sample, the corresponding false negatives at the \emph{RT} positions reported by the expert were identified as \emph{tentative true negatives} (TTN). Note that only FN with respective TTP were verified; in fact, more examples identified as FN may be correctly not detected by the system. False negatives that were not TTN were called semi-certain false negatives (FN).

Table \ref{Scan_results} reports the performance achieved on the testing set by the system with different CNN models. Additionally, the result intersection, i.e.\ detections consistent (in terms of \ label and \emph{RT}) among all the models, was evaluated. Results are presented in two forms: according to the expert-derived ground truth and according to the system-derived corrections. The total quantities of TP, TTP, FP (certain), FN (semi-certain) and TTN across all samples are given. We report the system sensitivity and mean average precision score (mAP), which is an evaluation metric commonly used in the object detection domain \cite{Everingham2010} (see Methods). Fig.\ \ref{fig:scan_figure2} shows a graphical example, in which TP, TTP, FN and TTN were identified. 

\textbf{\textsl{(2) -- VOC presence:}} 
To compute the system's specificity, we considered only the question of the presence of each VOC in each sample and ignore its \emph{RT} position.  
Since each VOC may appear in a sample at most once, such an approach leads to a binary classification problem. True negatives (TN) 
were thus well defined here as VOCs not detected in a sample by both the system and the expert-led processing. False positives were defined as VOCs detected by the system and not reported by the ground truth. In particular, this definition is different than above, i.e.,\ in (1) false positives covered also VOCs reported by both the system and the expert but at different \emph{RT} positions. 

\textsl{System-derived ground truth corrections:}
Similarly as above, false positives reported in their respective \emph{RT ranges} were identified as tentative true positives (TTP*), false positives that were not TTP were called certain false positives (FP*).

Table \ref{Scan_results2} reports the system specificity in the analysis of the 38 clinical samples from the testing set. Results are presented according to the expert-derived ground truth and according to the system-derived corrections. The total quantities of TN, FP* (certain), TTP* across all samples are given along with the system specificity.
The full tables of detections for each CNN model and their intersection, presented per each target VOC and per each testing sample, are reported in Supplementary Tables 3.1-6.2.

\begin{table*}[htbp]
\begin{center}
\caption{Evaluation (1) of the results of stage 2 analysis of testing GC-MS samples. Benchmarks: expert-derived ground truth (yellow) [tentative true positives (TTP) are considered FP, tentative true negatives (TTN) are considered FN]; system-derived correction (green) [tentative true positives (TTP) are considered TP, tentative true negatives (TTN) are considered TN].}

\begin{tabular}{P{1.5cm}|P{0.7cm}|P{0.73cm}|P{0.7cm}|P{0.73cm}|P{0.7cm}|p{2.2cm}|P{1.7cm}|P{1.6cm}|}
\cline{2-9}
                                                       & \multicolumn{5}{c|}{Classification}   &                                  \multirow{2}{*} {\textbf{ Benchmark}}                                         & \multicolumn{2}{c|}{Metrics}                                                                                                                                                              \\  \cline{1-6} \cline{8-9}
\multicolumn{1}{|c|}{Model}                            & \textbf{TP}           & \textbf{TTP}          & \textbf{FP}         & \textbf{TTN}          & \textbf{FN}         & 
& \textbf{sensitivity}                & \textbf{mAP}              \\ \hline
\multicolumn{1}{|l|}{}                                 &                       &                       &                     &                      &                      & \cellcolor[HTML]{FFFFE8}expert                                                               &  \cellcolor[HTML]{FFFFE8}0.9657 & \cellcolor[HTML]{FFFFE8}0.9019 \\ \cline{7-9} 
\multicolumn{1}{|l|}{\multirow{-2}{*}{\textbf{VGG-8-1D}}}       & \multirow{-2}{*}{816} & \multirow{-2}{*}{226} & \multirow{-2}{*}{2} & \multirow{-2}{*}{18} & \multirow{-2}{*}{11} & \cellcolor[HTML]{DEF3DE}\begin{tabular}[c]{@{}l@{}}system-derived \\ correction\end{tabular} &  \cellcolor[HTML]{DEF3DE}0.9896 & \cellcolor[HTML]{DEF3DE}0.9894 \\ \hline

\multicolumn{1}{|l|}{}                                 &                       &                       &                     &                      &                      & \cellcolor[HTML]{FFFFE8}expert                                                               & \cellcolor[HTML]{FFFFE8}0.9562 & \cellcolor[HTML]{FFFFE8}0.93 \\ \cline{7-9} 
\multicolumn{1}{|l|}{\multirow{-2}{*}{\textbf{DenseNet-40-1D}}} & \multirow{-2}{*}{808} & \multirow{-2}{*}{181} & \multirow{-2}{*}{6} & \multirow{-2}{*}{15} & \multirow{-2}{*}{22} & \cellcolor[HTML]{DEF3DE}\begin{tabular}[c]{@{}l@{}}system-derived \\ correction\end{tabular} &  \cellcolor[HTML]{DEF3DE}0.9782 & \cellcolor[HTML]{DEF3DE}0.9994 \\ \hline

\multicolumn{1}{|l|}{}                                 &                       &                       &                     &                      &                      & \cellcolor[HTML]{FFFFE8}expert                                                               & 0.9515 \cellcolor[HTML]{FFFFE8}  & \cellcolor[HTML]{FFFFE8}0.9282 \\ \cline{7-9} 
\multicolumn{1}{|l|}{\multirow{-2}{*}{\textbf{ResNet-34-1D}}} & \multirow{-2}{*}{804} & \multirow{-2}{*}{187} & \multirow{-2}{*}{9} & \multirow{-2}{*}{14} & \multirow{-2}{*}{27} & \cellcolor[HTML]{DEF3DE}\begin{tabular}[c]{@{}l@{}}system-derived \\ correction\end{tabular} &  \cellcolor[HTML]{DEF3DE}0.9735  & \cellcolor[HTML]{DEF3DE}0.998 \\ \hhline{|=|=|=|=|=|=|=|=|=|}

\multicolumn{1}{|l|}{}                                 &                       &                       &                     &                      &                      & \cellcolor[HTML]{FFFFE8}expert                                                               & 0.9325 \cellcolor[HTML]{FFFFE8}  & \cellcolor[HTML]{FFFFE8}0.9398 \\ \cline{7-9} 
\multicolumn{1}{|l|}{\multirow{-2}{*}{\textbf{Intersection}}} & \multirow{-2}{*}{787} & \multirow{-2}{*}{138} & \multirow{-2}{*}{1} & \multirow{-2}{*}{8} & \multirow{-2}{*}{49} & \cellcolor[HTML]{DEF3DE}\begin{tabular}[c]{@{}l@{}}system-derived \\ correction\end{tabular} &  \cellcolor[HTML]{DEF3DE}0.9497  & \cellcolor[HTML]{DEF3DE}1 \\ \hline

\end{tabular} 
\label{Scan_results}
\end{center}
\end{table*}

\begin{table*}[htbp]
\caption{Evaluation (2) of the results of stage 2 analysis of testing GC-MS samples. Benchmarks: expert-derived ground truth (yellow) [tentative true positives (TTP*) are considered FP*]; system-derived correction (green) [tentative true positives (TTP*) are considered TP].}
\begin{center}
\begin{tabular}{P{1.5cm}|P{0.7cm}|P{0.7cm}|P{0.9cm}|p{2.2cm}|P{1.7cm}|}
\cline{2-6}
& \multicolumn{3}{c|}{Classification} & \multirow{2}{*}{\textbf{ Benchmark}} & Metric \\\cline{1-4} \cline{6-6}
\multicolumn{1}{|c|}{Model}                   & \textbf{TN}           & \textbf{FP*}         & \textbf{TTP*}          & 
& \textbf{specificity}           \\ \hline
\multicolumn{1}{|l|}{}                                 &                       &                     &                       & \cellcolor[HTML]{FFFFE8}expert                                                              & \cellcolor[HTML]{FFFFE8}0.2915 \\ \cline{5-6} 
\multicolumn{1}{|l|}{\multirow{-2}{*}{\textbf{VGG-8-1D}}}       & \multirow{-2}{*}{86}  & \multirow{-2}{*}{1} & \multirow{-2}{*}{208} & \cellcolor[HTML]{DEF3DE}\begin{tabular}[c]{@{}l@{}}system-derived\\ correction\end{tabular} & \cellcolor[HTML]{DEF3DE}0.9885 \\ \hline
\multicolumn{1}{|l|}{}                                 &                       &                     &                       & \cellcolor[HTML]{FFFFE8}expert                                                              & \cellcolor[HTML]{FFFFE8}0.4237 \\ \cline{5-6} 
\multicolumn{1}{|l|}{\multirow{-2}{*}{\textbf{DenseNet-40-1D}}} & \multirow{-2}{*}{125} & \multirow{-2}{*}{4} & \multirow{-2}{*}{166} & \cellcolor[HTML]{DEF3DE}\begin{tabular}[c]{@{}l@{}}system-derived\\ correction\end{tabular} & \cellcolor[HTML]{DEF3DE}0.969 \\ \hline
\multicolumn{1}{|l|}{}                                 &                       &                     &                       & \cellcolor[HTML]{FFFFE8}expert                                                              & \cellcolor[HTML]{FFFFE8}0.3932 \\ \cline{5-6} 
\multicolumn{1}{|l|}{\multirow{-2}{*}{\textbf{ResNet-34-1D}}} & \multirow{-2}{*}{116} & \multirow{-2}{*}{6} & \multirow{-2}{*}{173} & \cellcolor[HTML]{DEF3DE}\begin{tabular}[c]{@{}l@{}}system-derived\\ correction\end{tabular} & \cellcolor[HTML]{DEF3DE}0.9508 \\ \hhline{|=|=|=|=|=|=|}

\multicolumn{1}{|l|}{}                                 &                       &                     &                       & \cellcolor[HTML]{FFFFE8}expert                                                              & \cellcolor[HTML]{FFFFE8}0.5593 \\ \cline{5-6} 
\multicolumn{1}{|l|}{\multirow{-2}{*}{\textbf{Intersection}}} & \multirow{-2}{*}{165} & \multirow{-2}{*}{0} & \multirow{-2}{*}{130} & \cellcolor[HTML]{DEF3DE}\begin{tabular}[c]{@{}l@{}}system-derived\\ correction\end{tabular} & \cellcolor[HTML]{DEF3DE}1 \\ \hline

\end{tabular}
\label{Scan_results2}
\end{center}
\end{table*}

\begin{table*}[htbp]
\end{table*}

\section*{Discussion} 

The results show that all tested system configurations, with various CNN models employed, achieved high performance in the detection of target VOCs in the clinical samples, reporting high sensitivity and specificity (when admitting the noisy-labelled ground truth). The results demonstrate that ion patterns can be effectively learnt directly from the raw GC-MS data. 
The problem of VOC detection includes various challenges such as detecting VOCs of low intensities and distinguishing overlapping VOCs and those with similar ion patterns (Table \ref{tabComposition}). Nevertheless, the system performed comparably well for all 30 VOCs from the dataset
(Supplementary Tables 3.1, 4.1, 5.1). Consequently, we claim that the CNN-based system proposed here allows bypassing time-consuming and labour-intensive expert-led data processing in real-world GC-MS data targeted analysis. 

We found strong evidence that
the proposed CNN-based system may outperform human experts. In our tests, the system detected 17\% to 23\% more occurrences of the VOCs than expert-led deconvolution-based method (TTP, Table \ref{Scan_results}). As much as 138 TTP were reported by each of the tested models. Additionally, 
the system did reveal possible errors, i.e.,\ incorrectly labelled VOCs (TTN). There are two possible reasons to explain the CNN-based system outperformance over the current expert-led processing. The current method involves complex preprocessing steps, based on the operator-subjective spectral deconvolution \cite{Coombes2007}. The proposed system instead analyses raw GC-MS data and thus bypasses multiple steps and possibly suboptimal choices of parameters. Another reason why the proposed workflow delivers more VOC detections may be its potential to detect compounds of low intensities. 
Additional analysis of the results (Supplementary Tables 1, 2.1, 3.1, 4.1) showed indeed that over 50\% of TTP (for VGG-8-1D; over 40\% for rest of the models) were reported by 25\% of target VOCs with the lowest average concentration (Table \ref{tabComposition}). Most TTP, 27, were reported for Propanoic acid (VOC 6), a compound of low concentration. Interestingly, before augmentation, the VOC dataset had only 22 training examples of VOC 6. Despite that, the CNN-based system was able to detect this compound correctly: none FP and FN reported, all 27 TTP reported by all the models. In our tests, the proposed system improves on the state-of-the-art performance of the current deconvolution-based process.

The proposed CNN-based system requires expert knowledge to be trained (stage 1), but consequently, it can detect VOCs autonomously and significantly faster than a human-driven procedure. The training stage in the proposed approach, requiring about 23 to 80 hours (depending on the CNN model), has to be performed only once to be able to scan new samples at a rate of as low as just around 2 minutes per sample (see Table \ref{tab:time} in Methods). Interestingly, the fastest among the tested architectures, i.e., VGG-8-1D, is also the one with the highest sensitivity. Consequently, the proposed system may support experts in much faster validation of hypotheses regarding compounds related to specific health conditions on new GC-MS samples. What is more, those hypotheses may be more accurate due to the system's ability to provide a more comprehensive list of VOCs than the deconvolution-based methods. 

Very deep networks are often thought to have more discrimination capabilities than shallower ones, but in this application, the VGG-like networks with reduced depth performed no worse than other very deep models (Table \ref{CVresults}). This suggests that the detection and classification of VOC ion patterns can be effectively performed with modest-depth networks, which are also less resource-intensive. All tested networks achieved higher performance when implemented with 1D filters, adapted to the nature of GC-MS data.

The proposed CNN-based approach provides specific information on individual VOCs in breath for further analysis and diagnosis, rather than a high-level sample classification (e.g.,\ \cite{VanBerkel2010}, \cite{Altomare2012}, \cite{Baranska2013}). 
As opposed to black-box and end-to-end machine learning diagnostic systems, the proposed system quickly produces accurate lists of VOCs, thus enabling a transparent and explainable pipeline for breathomics-based diagnosis.
This approach avoids the common problem of having limited datasets of clinical samples: each of many target VOC occurrences in a sample makes a data point; in this study, 82 clinical samples and 30 target VOCs resulted in a dataset with 3,736 unique data points and 373,600 augmented data points.

The proposed system does not use \emph{RT} values, which are a major source of variation among measurements with different instruments (or even the same instrument but on different days). Instead, the proposed system analyses the patterns of the VOCs, which remain substantially the same when are processed with similar instruments, in terms of type and properties such as resolution \cite{Garcia2011}. Therefore, the CNN-based system may be transferable to data obtained from another GC-MS instrument than the one from which the training dataset originates. Further tests are required to assess such proprietary precisely.

The proposed approach exploits GC-MS general properties and thus potentially extends beyond breath data. Further studies may extend tests to GC-MS data from a large variety of domains,\ e.g.: detection of CBRN (chemical, biological, radiological, nuclear) biomarkers in breath, saliva and skin for casualty triage \cite{Toxitriage}; tracking organic pollutants in water for environment monitoring \cite{Moore1984}; detection of accelerants in fire debris for criminal forensics \cite{Keto1991}; detecting drug ingredients in urine samples for law enforcement or sports anti-doping analysis \cite{Lee2018,Tsivou2006}; analysis of chemical composition of the planets' atmosphere in astrochemistry \cite{Krasnopolsky1981}; as well as in chemical engineering \cite{Tekin2014}, food, beverage and perfume analysis \cite{Bianchi2007, Garruti2006, VanAsten2002} and medicine \cite{Phillips2010}.

The proposed new approach to GC-MS data analysis, exploiting the application of deep learning, has the potential for extensive development in the future. Increasing breath analysis as a diagnostic technology will also increase the number of available GC-MS datasets: a larger number of VOC patterns, reflecting more of the possible variations in the data points, may benefit the accuracy of deep neural network training.
The use of GPU computing and dedicated hardware can help process the large amount of data collected through GC-MS; additionally, its rapid development, seen in recent years, may reduce the processing time even further. 

Future studies can extend the proposed CNN-based approach to also measure VOC intensities. The proposed system is currently limited to detecting the presence of VOCs of interest, but not their abundances. However, in real life scenarios, additional analysis by experts to determine peaks intensities and then concentrations of particular VOCs, e.g.,\ biomarkers related to a specific disease, may become necessary only if the system reveals their presence in a sample. In fact, as a result, the proposed workflow delivers along with a list of detected VOCs their \emph{RT} positions in the sample, which can significantly facilitate and accelerate the quantification of the compounds.

In summary, the proposed CNN-based system delivers a faster, more accurate and scalable method for automated targeted analysis of raw GC-MS data than the current state-of-the-art expert-led processing. The proposed approach has a significant potential to contribute to the development of breath analysis as a diagnostic platform to detect various diseases quickly, efficiently, and reliably.

\section*{Methods}

\subsection*{Ethical approval}
Ethical approval was sought and approved by the Ethics Committee of Edinburgh General Western Hospital to obtain the clinical breath samples (REC reference: 16/SS/0059, and IRAS project ID: 19961).

\subsection*{Breath sample collection}
Breath samples were collected from 25 participants before and after radiotherapy at 1, 3, and 6 hr. A Respiration Collector for In Vitro Analysis ReCIVA$\texttrademark$  device (Owlstone Medical, Cambridge, UK) was used for the breath samples collection. Clean air was provided from room-air filtered with an activated-carbon scrubber and HEPA filter, at a flow of 35 dm$^3$ min$^{-1}$. (The air-supply unit was built and tested by the Centre for Analytical Science, Loughborough, UK). The total sample volume of breath was set to 1000 cm$^3$ with a sampling duration cap of 900 s. Tenax\textsuperscript{\textregistered}/Carbotrap 1TD hydrophobic adsorbent tube (Markes International Ltd, Llantrisant, UK) was used. All materials were conditioned and sterilised before use to reduce exogenous VOC artefacts. clinical staff were trained and proficiency-tested prior to clinical breath sampling. Environmental and air-supply samples were collected with each set of breath samples. Samples were sealed and stored at ca. 4\textdegree{}C immediately after collection and transported to Loughborough Centre for Analytical Science within 48 hr \cite{Thomas}. Samples were dry-purged on receipt with a 120 cm$^3$ of purified nitrogen at a flow rate of 60 cm$^3$ min$^{-1}$. Toluene-D8 (0.069 ng) and trichloromethane-d (0.28 ng) internal standards were spiked into the sample during the dry-purge process using a six-port valve. All samples were then sealed and stored at 4\textdegree{}C prior to analysis.

\subsection*{GC-MS processing of clinical samples}
Thermal desorption (Unity-2, Markes International) interfaced to a GC (Agilent, 7890A) coupled to a quadrupole mass spectrometer (Agilent, MS 5977A) was used for the analysis of all clinical samples, see Supplementary Table 7 for operating details. The ion channels from 40 \emph{m/z} to 450 \emph{m/z} were measured with unit resolution. The instrumental scanning rate was approximately 6.25 Hz, which for about one-hour processing gave approximately 22500 \emph{RT} points. The size of the derived abundance matrix of each sample is $R \times 411$, where $R\approx22500$. 

The samples were analysed over a year period (Sep 2016 - Sep 2017) as part of a wider multi-centre clinical study campaign. Instruments were serviced whenever statistical process controls indicated a z-score $>3$ or more than 3 consecutive z-scores $>2$. The frequency of the service  interventions was determined by the quality and contamination levels of the samples returned from the clinics. The column was replaced once during this phase of the campaign, in March 2017.

\subsection*{Abundance matrix}
For each sample, GC-MS processing produces an abundance matrix $A\in\mathbb{R}^{\emph{R}\times 411}$. Let $z_{i}\in\mathbb{R}^{411}$, $i=1, ..., R$ be mass spectrum (i.e., intensities across consecutive ion channels derived by MS) at particular retention time points $r_i$, i.e.:

\[
A=\left[\begin{array}{c}
z_{1}\\
\vdots\\
z_{R}
\end{array}\right].
\]

\noindent For each ion channel $z_i$ its corresponding \emph{RT} point $r_i$ is specified by the function $\mathcal{RT}$:

\[
\mathcal{RT}:\;z_{i}\rightarrow r_{i}\in\mathbb{R}^{+}.
\]

\subsection*{Expert-led GC-MS data processing}

GC-MS data denoising and baseline correction and feature deconvolution were carried out (AnalyzerPro Spectral Works, UK) and 350 to 500 VOC features per sample were recorded. Features were aligned to correct for retention time variation, and the VOCCluster algorithm  \cite{Alkhalifah2019} was used to cluster all features into groups. Each feature was assigned an identifier in the format: BRI - $m/z_1$ $m/z_2$ $...$ $m/z_n$. BRI indicated the retention index for the VOC breath feature and $m/z_1 \, ... \, m/z_n$ are the nominal masses of the ion-fragments in decreasing order of abundance needed to uniquely define the feature. The processing time of a single breath sample is estimated as 60 to 120 minutes.

To produce the ground truth for this study, this process delivered a list of VOCs with Level 2 chemical identification \cite{Salek2013} and their corresponding positions in the matrix of raw data, this is $startRT$, $peakRT$ (denoted also as \emph{RT}) and $endRT$ - the indexes along the retention time where the compound was measured to start, peak and end the release from the GC column.

The process was not completed for 5 breath samples for technical issues, resulting in 120 expert-processed GC-MS samples.

\subsection*{VOC dataset structure}

The VOC dataset was extracted from the abundance matrices of the 82 samples from the training set, using the ground truth for the 30 target VOCs. A VOC data point $\hat{s}$ has a size 411 along the \emph{m/z} dimension, which ensures inclusion of the entire mass spectrum. The size along \emph{RT} dimension, $\delta$, was computed with the aim to capture the entire elution process for each target VOC instance. Precisely, the maximum duration of compound elution, \(max(endRT - startRT)\), was measured across 1868 occurrences of target VOCs reported by expert led-processing in training samples. This was computed as $\sim9.8$ seconds, corresponding to $\sim61$ time steps on the \emph{RT} axis. To allow for translation-based data augmentation, the size $\delta$ was increased by 19, resulting in a final $\delta=80$. Each data point $\hat{s}\in\mathbb{R}^{\delta\times 411}$ contains mass spectra of a VOC peak centred over \emph{RT} dimension, i.e.:

\[
\hat{s}=\left[z_{k},\ldots,\boldsymbol{z_{k+\frac{\delta}{2}-l}},...,\boldsymbol{z_{k+\frac{\delta}{2}}},...,\boldsymbol{z_{k+\frac{\delta}{2}+l}},\ldots,z_{k+\delta}\right]^T
\]
\noindent where

\[
\begin{array}{c}
\mathcal{RT}\left(z_{k+\frac{\delta}{2}-l}\right)=startRT,\\
\mathcal{RT}\left(z_{k+\frac{\delta}{2}+l}\right)=endRT,\\
\mathcal{RT}\left(z_{k+\frac{\delta}{2}}\right)=\mu,
\end{array}
\]

\noindent for $\mu=\frac{startRT+endRT}{2}$ -- a middle point of the VOC peak shape. Note that $\mu$ does not necessarily coincide with $peakRT$ value as VOC peaks can be not symmetric but skewed. $l$ value depends on the VOC instance.

The VOC dataset for CNN training was unbalanced: the groups representing target VOCs were unequal since each of the considered VOCs did not necessarily occur in each of the GC-MS samples from the training set. What is more, as during the scanning of entire raw breath samples the negative examples appear much often than target VOCs and cover a broad variety of ion patterns, the negative class was created as half of the VOC dataset. This aligns with common practice in object detection domain, where the negative class usually dominates with the ratio up to 3:1 \cite{Shrivastava2016, Liu2016}. The size of each class is reported in Supplementary Table 1.

\subsection*{Data augmentation}

In the proposed approach, we devised and tested two augmentation methods that maintain the underlying structure of a VOC pattern.

\textbf{\textsl{Translation along RT:}}
The VOC data point $\hat{s}$ is centred on the VOC peak so that the middle point of the VOC peak shape, $\mu=\frac{startRT+endRT}{2}$, is located at $\delta/2$. Twenty VOC data points were created from $\hat{s}$ by shifting the extraction point from the abundance matrix $A$ from -9 to +10 steps along the \emph{RT} axis (0 indicates $\hat{s}$), i.e.:

\[
\hat{s}_{n}^{*}=\left[z_{k+n},\ldots,\boldsymbol{z_{k+\frac{\delta}{2}}},\ldots,z_{k+n+\frac{\delta}{2}},\;\;\;\ldots\;\;\;,z_{k+n+\delta}\right]^T
\]

\noindent Such data augmentation presents the VOC pattern at different positions in the data point: top, centre, bottom (translation along \emph{RT}) and represents variations in the pattern's location in the subsequent sub-matrices of $A$ seen by the network, while scanning entire raw GC-MS breath samples.

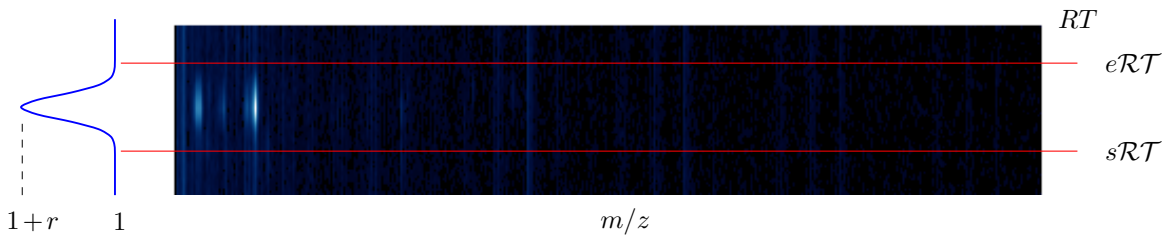
\begin{figure*}
\centering
\begin{tikzpicture}
   \node[anchor=south west,inner sep=0] (image) at (0,0) {
    \centerline{\includegraphics[width=0.75\linewidth]{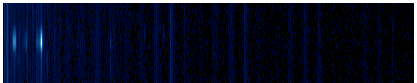}}};
    \begin{scope}[x={(image.south east)},y={(image.north west)}]
    \draw[line width=0.25mm,domain=0:1,smooth,variable=\s,blue]  plot ({0.08-0.08*exp(-((\s-0.5)^2)/0.009)},{\s});
    \draw [-, thin, red] (0.085, 0.25) -- (0.9, 0.25);
    \draw [-, thin, red] (0.085, 0.75) -- (0.9, 0.75);
    \node[text width=0.2cm] at (0.93,0.25) {$s\mathcal{RT}$};
    \node[text width=0.2cm] at (0.93,0.75) {$e\mathcal{RT}$};
    \node[text width=0.2cm] at (0.89,1) {$RT$};
    \node[text width=0.2cm] at (0.5,-0.15) {$m/z$};
    \node[text width=0.2cm] at (0.085,-0.15) {$1$};
    \node[text width=0.7cm] at (0.01,-0.15) {$1+r$};
    \draw [dashed, ultra thin, black] (0.001, 0) -- (0.001, 0.45);
    
    \end{scope}
\end{tikzpicture}
\caption{Scheme of the data augmentation by the intensity variation in VOC the data point. Along \emph{RT} points containing the VOC pattern (i.e.,\ between $s\mathcal{RT}$ and $e\mathcal{RT}$), the intensities of each \emph{m/z} channel were multiplied by a Gaussian-shaped function to simulate variations of the VOC concentration.
}
\label{fig:gauss}
\end{figure*}

\textbf{\textsl{Intensity variation:}}
In the VOC data point, the intensities along \emph{RT} interval corresponding to the VOC pattern were varied to simulate variations of the VOC concentration (see Fig.\ \ref{fig:gauss}). Background (i.e.,\ values along \emph{RT} not containing the pattern) remains unchanged. In this process, it was important to maintain the VOC ion pattern, i.e.,\ relative ratios along the $m/z$ dimension while increasing peak intensity along the \emph{RT} axis. This was achieved by multiplication of each mass spectrum $z_{startRT}, ..., z_{endRT}$ (corresponding to VOC location) by a particular value of a Gaussian-shaped function along $RT$ interval. Precisely, for a data point $\hat{s}$:

\[
\hat{s}=\left[\begin{array}{c}
z_{k}\\
\vdots\\
z_{k+\delta}
\end{array}\right].
\]

\noindent the augmented data point $\hat{s}^*$ can be computed as 

\[
\hat{s}^{*}=\left[\begin{array}{c}
z_{k}\cdot G_{\mathcal{RT}\left(z_{k}\right)}\\
\vdots\\
z_{k+\delta}\cdot G_{\mathcal{RT}\left(z_{k+\delta}\right)}
\end{array}\right],
\]

\noindent where

\[
G_{x}=\begin{cases}
e^{-\frac{1}{2}\left(\frac{x-\mu}{\sigma}\right)^{2}}\cdot r+1, & startRT<x<endRT\\
1, & \textrm{otherwise},
\end{cases}
\]

\noindent and $r$ is a random value in the range $(0, 0.1)$. 
This augmentation step was repeated to obtain 4 additional data points for each data point derived with translation along \emph{RT}. Therefore, the VOC dataset for CNN training was augmented 100 times (i.e., $20\times4$ times). The intensity values in each data point from the augmented dataset were normalised in the range \([0, 1]\).

The original VOC dataset had 3,736 data points. Following \emph{translation along RT} the dataset for CNN training consisted of 74,720 segments (partially-augmented dataset). When \emph{intensity variation} was also applied, the dataset contained 373,600 data points (fully-augmented dataset). Results presented in the Table \ref{CVresults} indicate that both developed augmentation methods bring benefits to the system: the best performance was achieved on the fully-augmented dataset.

\subsection*{CNN filter adaptation}
To adjust the deep learning models specifically to VOC detection, we adapt the CNN filters to the GC-MS data characteristics. In CNNs, filters specify the local receptive fields, i.e.,\ the regions of the image (or feature maps) visible by convolutional and pooling layers of the network at a time; this is an essential concept that enables CNNs to capture local geometric spatial correlations in the data \cite{LeCun1989}. With GC-MS data, such a local correlation occurs only in the retention time dimension. Along this dimension, the abundance of different \emph{m/z} increases and decreases depicting peaks as the VOC exit the GC column. On the other hand, the abundance values across different \emph{m/z} channels also correlate as the particular ions make up the VOC pattern (Fig.\ \ref{fig:breathogram}a,b). However the values along this dimension represent independent ion channels and locally they are only weakly correlated, thus their correlation cannot be captured by small local filters.

One of the hypotheses in this study is that the convolutional and pooling layers in the network do not need to be two-dimensional as it is usual for image classification. Thus, two types of filters are tested: a traditional two-dimensional filter and a specific one-dimensional filter along the \emph{RT} axis to cover only this dimension.
The filters sizes in the particular network layers are given, along with the detailed architecture of each network, in Supplementary Tables 2.1-2.11.

\subsection*{Phase A output}
The last layer in all the tested networks is a fully connected layer with softmax activation. The softmax function $\sigma:\mathbb{R}^{k}\rightarrow\mathbb{R}^{k}$, defined as

\[
\sigma(\mathbf{x})_{i}=\frac{e^{x_{i}}}{{\displaystyle \sum}_{j=0}^{k}e^{x_{j}}}
\]

\noindent for $\;i=1, ..., k$, $\;\mathbf{x}=\left(x_{i}, ..., x_{k}\right)\in\mathbb{R}^{k}$,
takes as input a vector of $k$ real numbers and normalises it into a probability distribution consisting of $k$ probabilities; all the components are mapped to the interval $(0,1)$ such that their sum is 1. Therefore, the network returns a probability distribution over output classes, i.e the probabilities of the allocation of each particular normalised data point $s\in\mathbb{R}^{\delta\times 411}$ to each of 31 VOC groups: 

\[
s\rightarrow\left(P_{0}(s), ..., P_{30}(s)\mid P_{i}(s)=P\left(s\in C_{i}\right)\right)
\]

\noindent for $i=0,\ldots,30$, $C_i$ - class representing the VOC labelled with $i$. (Note that $P_i(s)=\sigma(\mathbf{x})_{i}$, where $\mathbf{x}$ is a vector of 31 entries generated for data point $s$ by the last fully connected layer of the network.)

The VOC class that gives the highest probability for a data point $s$ is selected as its classification $c$:

\[
c(s)=\arg\max_{i}\left\{ P_{i}(s)\mid i=0, ..., 30\right\}
\]

\noindent with confidence $t$:

\[
t(s)=\max_{i}\left\{ P_{i}(s)\mid i=0, ..., 30\right\}.
\]

\noindent As a result, for each analysed raw GC-MS sample with abundance matrix $A\in\mathbb{R}^{\mathrm{R}\times 411}$, the CNN scanning in phase A produces two sequences $L_A$ and $T_A$. $L_A$ values indicate classification labels of the subsequent sub-matrices $s$ of $A$:

\[
L_A=\left(c(s_{i})\mid s_{i}\subset A,\;i=1, ..., N\right);
\]

\noindent $T_A$ values indicate classification confidence given by the network for each respective label:

\[
T_A=\left(t(s_{i})\mid s_{i}\subset A,\;i=1, ..., N\right),
\]

\noindent where $N=RT-\delta+1\approx22500$ is a number of data points in $A$.

\subsection*{Duration rule: VOC detection} 

Each maximal sub-sequence of $L_A$ with constant label values $j$ and length equal or greater than a constant $\gamma$ indicates one detection $d_j$ of the VOC $j$, i.e.:

\[
d_j=(s_{l}, ..., s_{l+n-1}),
\]
where
\[
c\left(s_{i}\right)=j, \; i\in\left[l,l+n-1\right],
\; \; \;
c\left(s_{l-1}\right)\neq j,\; c\left(s_{l+n}\right)\neq j,
\; \; \;
n\geq\gamma, \; j>0.
\]

\noindent The value $\gamma$ was derived from the width setting of matrix $s$: for the augmentation, width of $s$ was enlarged by 19 pixels in respect to the maximum peak width (see Data augmentation).
Hence, during the scanning of the abundance matrix $A$, the entire shape of a target VOC peak is seen by the model at least $\gamma=20$ times.
All the detections, of any class, from the abundance matrix $A$ (i.e., sample) constitutes the set $\mathcal{D}_A$.

\subsection*{RT of detection}
\noindent Let the retention time value corresponding to the specific data point $s=\left[z_{k},\ldots,z_{k+\delta}\right]$ be defined as the retention time value of its middle ion channel, i.e.: 

\[
\mathcal{RT}\left(s\right)=\mathcal{RT}\left(z_{k+\frac{\delta}{2}}\right).
\]

\noindent Retention time values corresponding to the first and last data points $s_i$ of detection $d_{j}$ are called respectively the \emph{start detection}, 
$s\mathcal{RT}(d_j)$, and \emph{end detection}, $e\mathcal{RT}(d_j)$:

\[
s\mathcal{RT}\left(d_{j}\right)=\min\left\{RT\left(s_{i}\right)\mid s_{i}\in d_{i}\right\},
\]

\[
e\mathcal{RT}\left(d_{j}\right)=\max\left\{RT\left(s_{i}\right)\mid s_{i}\in d_{i}\right\}.
\]

\noindent We denote the resulting detection interval by

\[
DI\left(d_{j}\right)=\left[s\mathcal{RT}\left(d_{j}\right),\:e\mathcal{RT}\left(d_{j}\right)\right].
\]

\subsection*{Order rule} 
The elution of particular VOCs from the GC column can be expected in a certain order (small variations may occur) 
related to compounds' chemical properties. Therefore, to derive a reliable list of detected VOCs, the detections that fall outside the expected order are removed.

Let $\leq$ be a linear order on a set $\mathcal{D}_A$ of detections, such that 

\[
 \forall d_{i},d_{j}\in\mathcal{D}_A:\;d_{i}\leq d_{j}\Longleftrightarrow s\mathcal{RT}\left(d_{i}\right)\leq s\mathcal{RT}\left(d_{j}\right).
\]

\noindent Then:

\begin{equation*}
\begin{split}
\tilde{\mathcal{D}}_A=\mathcal{D}_A\setminus\{d_{f} & \mid \exists\, i,j,k<f: d_f<d_i<d_j<d_k \\
& \vee\exists\, i,j,k>f: d_i<d_j<d_k<d_f \} .
\end{split}
\end{equation*}

\subsection*{Detection confidence} For each detection $d_j$, the detection confidence $\mathcal{T}(d_j)$ was calculated as the maximum value of the moving average, with size $\gamma=20$, of the classification confidence values $t(s)$ for the consecutive data points $s$ from $d_j$, i.e.:

\begin{equation*} 
\mathcal{T}\left(d_{j}\right)=\max_{\left(s_{i}, ..., s_{i+\gamma-1}\right)\subset d_{j}}\left\{ \frac{t\left(s_{i}\right)+...+t\left(s_{i+\gamma-1}\right)}{\gamma}\right\}.
\end{equation*}

\noindent The value $\gamma=20$ was derived as explained for the duration rule above.

\subsection*{Uniqueness rule} 
Since each VOC may be present in a GC-MS sample at most once, if multiple detections of one VOC occur, $d_{j}^{1}, ..., d_{j}^{K_j}$, the one with the highest detection confidence value is kept in the set of detections:

\[
d_{j}(A)=\arg\max_{d_{j}^{k}\in \mathcal{D}_A}\left\{ \mathcal{T}\left(d_{j}^{k}\right)\mid k=1, ..., K_j\right\},
\]
\[
\dbtilde{\mathcal{D}}_A=\mathcal{D}_A\setminus\left\{d_{j}^{k}\neq d_{j}(A)\mid k=1, ..., K_j, \; j=1, ..., 30\right\}.
\]

\subsection*{Phase B output}
As an output, phase B produces a set $\mathcal{L}$ of detections $d_j \in \dbtilde{\mathcal{D}}_A$ as 4-tuples of label, start detection, end detection and detection confidence:

\begin{equation*}
\begin{split}
\mathcal{L}= \{
& \left[\begin{array}{cccc}
 j, & s\mathcal{RT}\left(d_{j}\right), & e\mathcal{RT}\left(d_{j}\right), & \mathcal{T}(d_{j})\end{array}\right] : \\
&  d_j \in \dbtilde{\mathcal{D}}_A,\; j=1, ..., 30\}.
\end{split}
\end{equation*}

\subsection*{RT range of a VOC}
\noindent 
Each target VOC in the GC-MS dataset is narrowly distributed over a specific range of the \emph{RT} dimension. From the processed data files, the \emph{RT} range of occurrences was extracted for each target VOC $j$ across all its instances in the GC-MS dataset reported by expert-led processing, i.e.:

\[
RTrange_j = \left[\min\{peakRT_j\}, \max\{peakRT_j\}\right].
\] 

\noindent What is important, such \emph{RT} ranges depend on the GC column. During the course of our study, the GC column was changed and thus we extracted two \emph{RT} ranges for each VOC, $RTrange_j^1$ and $RTrange_j^2$, valid respectively before and after column change. Supplementary Table 1 presents \emph{RT} ranges for each target VOC.

\subsection*{Tentative true positives}
Let $\hat{d}_{j}$ be the detections in the set $\mathcal{D}_A$
that were not reported in the ground truth at the \emph{RT} position reported by the system for that sample $A$. Note that we consider the set $\mathcal{D}_A$, i.e., before reducing the number of detections for the order and uniqueness rules, as their application changes the detection distribution. We observed that most (60\% to 74\% depending on the network) detections 
$\hat{d}_{j}\in\bigcup_{A}\mathcal{D}_A$ were reported by the system in the \emph{RT} range of the $j$ VOC (accordingly before or after column change, $i=1$ or $2$), i.e.:

\[
DI\left(\hat{d_{j}}\right) \cap RTrange_{j}^{i}\neq\emptyset,
\]

\noindent Let $I$ denote an interval of length 

\[
\left|I\right|\leq\max_{\hat{d_{j}}\in\bigcup_{A}\mathcal{D}_A}\left|DI\left(\hat{d_{j}}\right)\right|,
\]

\noindent chosen uniformly at random from within the \emph{RT} dimension (of length $R$). Then the probability of the intersection of the interval $I$ with a given $RTrange_{j}^{i}$ can be computed as:

\begin{equation*}
\begin{split}
P & \left(I \cap RTrange_{j}^{i}\neq\emptyset\right)=\frac{\left|RTrange_{j}^{i}\right|+\left|I\right|}{R-\left|I\right|} \\
&  \leq \frac{\max_{i,j}\left|RTrange_{j}^{i}\right|+\max_{\hat{d_{j}}}\left|DI\left(\hat{d_{j}}\right)\right|}{R-\max_{\hat{d_{j}}}\left|DI\left(\hat{d_{j}}\right)\right|}\coloneqq P_{max},
\end{split}
\end{equation*}
where the $\max_{\hat{d_{j}}\in\bigcup_{A}\mathcal{D}_A}$ depends on the network used. $P_{max}$ is the upper bound probability of a random (false) detection $\hat{d}_{j}\in\mathcal{D}_A$ of a target VOC $j$ at specific \emph{RT} range in the sample $A$. We computed the following bounds:

\begin{table}[!htbp]
\begin{center}
\begin{tabular}{P{2.7cm}|P{2.7cm}}
    Network & $P_{max}$ \\
    \hline
    VGG-8-1D & 0.0431 \\
    DenseNet-40-1D & 0.0439 \\
    ResNet-38-2D & 0.0441
\end{tabular}
\end{center}
\end{table}

\noindent The probabilities $P_{max}$ are all very low in comparison to the actual proportion of such considered detections (i.e., 60\% to 74\% depending on the network). Therefore, it is highly likely that the detections $\hat{d}_{j}$ within a compatible \emph{RT} range are not random false detections, but are correctly detected VOCs. 
Such detections are named \emph{tentative true positives} (TTP).

\subsection*{Tentative true negatives}
Several VOCs, not detected at the specific \emph{RT} points reported by the ground truth (preliminary marked as false negatives), were detected as tentative true positives on the different \emph{RT} positions within their compatible \emph{RT} range. Because of the high chance of tentative true positives to be true positives, the noisy-labelled character of the ground truth and the constraints that a VOC appears only once in a sample, we conclude that such particular examples may indicate errors in the expert-led processing. Such examples are called \emph{tentative true negatives}.

\subsection*{Average precision score} 
The system performance of each network was assessed by an average precision score (AP) for each target VOC and mean average precision (mAP). For each target VOC $j$, a list of all system detections $d_{j}\in\bigcup_{A}\dbtilde{\mathcal{D}}_A$ of the VOC in all clinical samples from the testing set was derived from stage 2 output and sorted in descending order associated with the detection confidence scores $\mathcal{T}\left(d_{j}\right)$ (for Intersection of the models, it was sorted by mean value of detection confidence scores for tested models).
For the first $n$ elements of this list, the precision function $Precision(n)$ was defined as the proportion of TP.
The recall function $Recall(n)$ was defined as the proportion of all detections in the ground truth that appear in the first $n$ elements of the system detection list. As usual, the AP score for each target VOC was calculated as the integral under the graph of precision against recall (Supplementary Figures 3.1-6.30). mAP was calculated as the average of AP values across all target VOCs. AP values for each target VOC for each tested model, along with the respective graphs of precision vs recall functions, are given in the Supplementary Information.

\subsection*{Sensitivity and specificity}
The system's sensitivity was computed forthe \emph{expert} benchmark (i.e., TTP are considered FP, TTN are considered FN) as

\[
sensitivity=\frac{TP}{TP+FN+TTN},
\]

\noindent and for the \emph{system-derived correction} benchmark (i.e., TTP are considered TP, TTN are considered TN) as

\[
sensitivity=\frac{TP+TTP}{TP+TTP+FN}.
\]

\noindent The system specificity was computed for the \emph{expert} benchmark (i.e., TTP are considered FP) as

\[
specificity=\frac{TN}{TN+FP*+TTP*},
\]

\noindent and for the \emph{system-derived correction} benchmark (i.e., TTP are considered TP) as

\[
specificity=\frac{TN}{TN+FP*}.
\]

\subsection*{Resources}
All experiments were run on a server running Linux Ubuntu with 20 cores, 128GB RAM and NVIDIA Tesla K80 GPU cards. Table \ref{tab:time} compares the training time (stage 1, Fig.\ \ref{fig:scheme}) and average scanning time (stage 2) of each of the tested network architectures. Memory requirements are given in the Supplementary Tables 2.1-2.11.

\begin{table}
\caption{Time resources: time of a network training on the fully-augmented VOC dataset and scanning time of a single GC-MS sample. \label{tab:time}}
\begin{center}
\begin{tabular}{|P{2.7cm}|P{2cm}|P{2cm}|}
    \hline
    Network & Training time & Scan time\\
    \hline
    VGG-8-1D & 23 h & 2 min\\
    DenseNet-40-1D & 37 h & 10 min\\ 
    ResNet-38-2D & 80 h & 12.5 min\\\hline
\end{tabular}
\end{center}
\end{table}


\section*{Contributions}
A.Sk. and A.So. designed the research with contributions from D.S., C.L.P.T. and M.S. A.Sk. designed and developed the proposed system, and run the computer experiments. D.S. analysed the clinical samples on GC-MS and carried out manual data processing. C.L.P.T. co-ordinated the cross-institute effort and secured funding. A.Sk, E.H, D.S., M.S. and A.So. carried out the result analysis.  A.Sk. and A.So. created the figures and wrote the paper. K.D. collected the clinical samples from subjects as Research Nurse working on the trial. M.E designed and oversaw the clinical trial with contributions from W.H.N and D.B.M. W.H.N and D.B.M. oversaw the radiotherapy arm of the study. All authors provided feedback and contributed to improving and finalising the paper.

\section*{Acknowledgements}
We are thankful to Iain Phillips and Yaser Al Khalifah for sharing their work on VOC clustering, Yang Hu for discussions on deep neural networks, Joanna Turner for discussions on the analogy with object detection. This study was partially funded by the EU H2020 TOXI-Triage Project \#653409.



\bibliographystyle{unsrt}  
\bibliography{references.bib} 

\begin{thebibliography}{10}

\bibitem{Smolinska2014}
A~Smolinska, A~C Hauschild, R~R Fijten, J~W Dallinga, J~Baumbach, and F~J van
  Schooten.
\newblock {Current breathomics-a review on data pre-processing techniques and
  machine learning in metabolomics breath analysis}.
\newblock {\em Journal of Breath Research}, 8(2):27105, 2014.

\bibitem{Hollywood2006}
Katherine Hollywood, Daniel~R. Brison, and Royston Goodacre.
\newblock {Metabolomics: Current technologies and future trends}, 2006.

\bibitem{Rattray2014}
Nicholas~J.W. Rattray, Zahra Hamrang, Drupad~K. Trivedi, Royston Goodacre, and
  Stephen~J. Fowler.
\newblock {Taking your breath away: Metabolomics breathes life in to
  personalized medicine}, 2014.

\bibitem{VanBerkel2010}
J.~J B~N {Van Berkel}, J.~W. Dallinga, G.~M. M{\"{o}}ller, R.~W~L Godschalk,
  E.~J. Moonen, E.~F~M Wouters, and F.~J. {Van Schooten}.
\newblock {A profile of volatile organic compounds in breath discriminates COPD
  patients from controls}.
\newblock {\em Respiratory Medicine}, 104(4):557--563, 2010.

\bibitem{Li2017}
Wenwen Li, Yong Liu, Yu~Liu, Shouquan Cheng, and Yixiang Duan.
\newblock {Exhaled isopropanol: new potential biomarker in diabetic breathomics
  and its metabolic correlations with acetone}.
\newblock {\em RSC Advances}, 7(28):17480--17488, 2017.

\bibitem{Fuchs2010}
Patricia Fuchs, Christian Loeseken, Jochen~K. Schubert, and Wolfram Miekisch.
\newblock {Breath gas aldehydes as biomarkers of lung cancer}.
\newblock {\em International Journal of Cancer}, 126(11):2663--2670, 2010.

\bibitem{Altomare2012}
D~F Altomare, M~{Di Lena}, F~Porcelli, L~Trizio, E~Travaglio, M~Tutino,
  S~Dragonieri, V~Memeo, and G~de~Gennaro.
\newblock {Exhaled volatile organic compounds identify patients with colorectal
  cancer}.
\newblock {\em British Journal of Surgery}, 100:144--150, 2012.

\bibitem{Phillips2010}
Michael Phillips, Renee~N Cataneo, Christobel Saunders, Peter Hope, Peter
  Schmitt, and James Wai.
\newblock {Volatile biomarkers in the breath of women with breast cancer.}
\newblock {\em Journal of breath research}, 4(2):026003, 2010.

\bibitem{Watson2008}
J.~Throck Watson and O.~David Sparkman.
\newblock {\em {Introduction to Mass Spectrometry: Instrumentation,
  Applications and Strategies for Data Interpretation: Fourth Edition}}.
\newblock 2008.

\bibitem{stein1999integrated}
Stephen~E Stein.
\newblock An integrated method for spectrum extraction and compound
  identification from gas chromatography/mass spectrometry data.
\newblock {\em Journal of the American Society for Mass Spectrometry},
  10(8):770--781, 1999.

\bibitem{Hubschmann2015}
Hans-Joachim H{\"{u}}bschmann.
\newblock {\em {Handbook of GC-MS}}.
\newblock 2015.

\bibitem{KovatsRI}
{The Kov{\'{a}}ts Retention Index System}.
\newblock {\em Analytical Chemistry}, 36(8):31A--41A, 2012.

\bibitem{Colby1992}
Bruce~N. Colby.
\newblock {Spectral deconvolution for overlapping GC/MS components}.
\newblock {\em Journal of the American Society for Mass Spectrometry},
  3(5):558--562, 1992.

\bibitem{Alkhalifah2019}
Yaser Alkhalifah, Iain Phillips, Andrea Soltoggio, Kareen Darnley, William~H.
  Nailon, Duncan McLaren, Michael Eddleston, Paul Thomas, and Dahlia Salman.
\newblock {VOCCluster: Untargeted Metabolomics Feature Clustering Approach for
  Clinical Breath Gas Chromatography - Mass Spectrometry Data}.
\newblock 2019.

\bibitem{Ren2015}
Sheng Ren, Anna~A. Hinzman, Emily~L. Kang, Rhonda~D. Szczesniak, and Long~Jason
  Lu.
\newblock {Computational and statistical analysis of metabolomics data}, 2015.

\bibitem{Coombes2007}
Kevin~R. Coombes, Keith~A. Baggerly, and Jeffrey~S. Morris.
\newblock {Pre-processing mass spectrometry data}.
\newblock In {\em Fundamentals of Data Mining in Genomics and Proteomics},
  pages 79--102. 2007.

\bibitem{Likic2009}
Vladimir~A. Liki{\'{c}}.
\newblock {Extraction of pure components from overlapped signals in gas
  chromatography-mass spectrometry (GC-MS)}, 2009.

\bibitem{PaulSajda2006}
{Paul Sajda}.
\newblock {Machine learning for detection and diagnosis of disease.}
\newblock {\em Annual review of biomedical engineering}, 8(April):8.1--8.29,
  2006.

\bibitem{Mamoshina2016}
Polina Mamoshina, Armando Vieira, Evgeny Putin, and Alex Zhavoronkov.
\newblock {Applications of Deep Learning in Biomedicine}, 2016.

\bibitem{Baranska2013}
Agnieszka Baranska, Ettje Tigchelaar, Agnieszka Smolinska, Jan~W. Dallinga,
  Edwin~J.C. Moonen, Jackie~A.M. Dekens, Cisca Wijmenga, Alexandra Zhernakova,
  and Frederik~J. {Van Schooten}.
\newblock {Profile of volatile organic compounds in exhaled breath changes as a
  result of gluten-free diet}.
\newblock {\em Journal of Breath Research}, 7(3), 2013.

\bibitem{LeCun1989}
Y.~LeCun, B.~Boser, J.~S. Denker, D.~Henderson, R.~E. Howard, W.~Hubbard, and
  L.~D. Jackel.
\newblock {Backpropagation Applied to Handwritten Zip Code Recognition}, 1989.

\bibitem{Rawat2017}
Waseem Rawat and Zenghui Wang.
\newblock {Deep convolutional neural networks for image classification: A
  comprehensive review}, 2017.

\bibitem{LeCun2004}
Yann LeCun, Fu~Jie Huang Fu~Jie Huang, and L.~Bottou.
\newblock {Learning Methods for Generic Object Recognition with Invariance to
  Pose and Lighting}.
\newblock {\em Computer Vision and Pattern Recognition, 2004. CVPR 2004.
  Proceedings of the 2004 IEEE Computer Society Conference on}, 2:II--97 --
  104, 2004.

\bibitem{Ciresan2012}
Dan Cireşan, Ueli Meier, and Juergen Schmidhuber.
\newblock {Multi-column Deep Neural Networks for Image Classification}.
\newblock {\em International Conference of Pattern Recognition},
  (February):3642--3649, 2012.

\bibitem{Krizhevsky2012}
Alex Krizhevsky, Ilya Sutskever, and Geoffrey~E Hinton.
\newblock {ImageNet Classification with Deep Convolutional Neural Networks}.
\newblock {\em Nips}, pages 1--9, 2012.

\bibitem{Sermanet2013}
Pierre Sermanet, David Eigen, Xiang Zhang, Michael Mathieu, Rob Fergus, and
  Yann LeCun.
\newblock {OverFeat: Integrated Recognition, Localization and Detection using
  Convolutional Networks}.
\newblock {\em arXiv preprint arXiv}, page 1312.6229, 2013.

\bibitem{Nielsen2015}
Michael~A. Nielsen.
\newblock {Neural Networks and Deep Learning}.
\newblock In {\em Machine Learning}, pages 875--936. 2015.

\bibitem{Garcia2011}
Antonia Garcia and Coral Barbas.
\newblock {Gas chromatography-mass spectrometry (GC-MS)-based metabolomics.}
\newblock {\em Methods in molecular biology (Clifton, N.J.)}, 708:191--204,
  2011.

\bibitem{Skarysz2018}
Angelika Skarysz, Yaser Alkhalifah, Kareen Darnley, Michael Eddleston, Yang Hu,
  Duncan~B. McLaren, William~H. Nailon, Dahlia Salman, Martin Sykora, C.~L.Paul
  Thomas, and Andrea Soltoggio.
\newblock {Convolutional neural networks for automated targeted analysis of raw
  gas chromatography-mass spectrometry data}.
\newblock In {\em Proceedings of the International Joint Conference on Neural
  Networks}, volume 2018-July, 2018.

\bibitem{Cortes1995}
Corinna Cortes and Vladimir Vapnik.
\newblock {Support-Vector Networks}.
\newblock {\em Machine Learning}, 20(3):273--297, 1995.

\bibitem{Zhang2000}
G.P. Zhang.
\newblock {Neural networks for classification: a survey}.
\newblock {\em IEEE Transactions on Systems, Man and Cybernetics, Part C
  (Applications and Reviews)}, 30(4):451--462, 2000.

\bibitem{Simonyan2014}
Karen Simonyan and Andrew Zisserman.
\newblock {Very Deep Convolutional Networks for Large-Scale Image Recognition}.
\newblock {\em arXiv preprint arXiv:1409.1556}, pages 1--13, 2014.

\bibitem{He2016}
Kaiming He, Xiangyu Zhang, Shaoqing Ren, and {Jian Sun}.
\newblock {Deep Residual Learning for Image Recognition}.
\newblock {\em Multimedia Tools and Applications}, pages 1--17, 2016.

\bibitem{huang2017densely}
Gao Huang, Zhuang Liu, Kilian~Q Weinberger, and Laurens van~der Maaten.
\newblock Densely connected convolutional networks.
\newblock In {\em Proceedings of the IEEE conference on computer vision and
  pattern recognition}, volume~1, page~3, 2017.

\bibitem{Dyk2001}
David~A.Van Dyk and Xiao~Li Meng.
\newblock {The art of data augmentation}.
\newblock {\em Journal of Computational and Graphical Statistics}, 10(1):1--50,
  2001.

\bibitem{Hastie2009}
Trevor Hastie, Robert Tibshirani, and Jerome Friedman.
\newblock {\em {The Elements of Statistical Learning (Second Edition, 10th
  print)}}, volume~1.
\newblock 2009.

\bibitem{Everingham2010}
Mark Everingham, Luc {Van Gool}, Christopher~K.I. Williams, John Winn, and
  Andrew Zisserman.
\newblock {The pascal visual object classes (VOC) challenge}.
\newblock {\em International Journal of Computer Vision}, 88(2):303--338, 2010.

\bibitem{Toxitriage}
{TOXI-triage project, http://toxi-triage.eu}.

\bibitem{Moore1984}
R.~A. Moore and F.~W. Karasek.
\newblock {GC/MS identification of organic pollutants in the caroni river,
  trinidad}.
\newblock {\em International Journal of Environmental Analytical Chemistry},
  17(3-4):203--221, 1984.

\bibitem{Keto1991}
Raymond~O. Keto and Philip~L. Wineman.
\newblock {Detection of Petroleum-Based Accelerants in Fire Debris by Target
  Compound Gas Chromatography/Mass Spectrometry}.
\newblock {\em Analytical Chemistry}, 63(18):1964--1971, 1991.

\bibitem{Lee2018}
Junhui Lee, Jiwon Park, Ahra Go, Heesung Moon, Sujin Kim, Sohee Jung, Wonjoon
  Jeong, and Heesun Chung.
\newblock {Urine Multi-drug Screening with GC-MS or LC-MS-MS Using SALLE-hybrid
  PPT/SPE}.
\newblock {\em Journal of Analytical Toxicology}, 42(9):617--624, 2018.

\bibitem{Tsivou2006}
M.~Tsivou, N.~Kioukia-Fougia, E.~Lyris, Y.~Aggelis, A.~Fragkaki, X.~Kiousi,
  Ph~Simitsek, H.~Dimopoulou, I.~P. Leontiou, M.~Stamou, M.~H. Spyridaki, and
  C.~Georgakopoulos.
\newblock {An overview of the doping control analysis during the Olympic Games
  of 2004 in Athens, Greece}.
\newblock {\em Analytica Chimica Acta}, 555(1):1--13, 2006.

\bibitem{Krasnopolsky1981}
V.~A. Krasnopolsky and V.~A. Parshev.
\newblock {Chemical composition of the atmosphere of Venus}.
\newblock {\em Nature}, 292(5824):610--613, 1981.

\bibitem{Tekin2014}
Kubilay Tekin, Selhan Karag{\"{o}}z, and Sema Bektaş.
\newblock {A review of hydrothermal biomass processing}, 2014.

\bibitem{Bianchi2007}
F.~Bianchi, M.~Careri, M.~Musci, and A.~Mangia.
\newblock {Fish and food safety: Determination of formaldehyde in 12 fish
  species by SPME extraction and GC-MS analysis}.
\newblock {\em Food Chemistry}, 100(3):1049--1053, 2007.

\bibitem{Garruti2006}
Deborah~S. Garruti, Maria Regina~B. Franco, Maria Aparecida~A.P. {Da Silva},
  Nat{\'{a}}lia~S. Janzantti, and Gisele~L. Alves.
\newblock {Assessment of aroma impact compounds in a cashew apple-based
  alcoholic beverage by GC-MS and GC-olfactometry}.
\newblock {\em LWT - Food Science and Technology}, 39(4):373--378, 2006.

\bibitem{VanAsten2002}
Arian {Van Asten}.
\newblock {The importance of GC and GC-MS in perfume analysis}.
\newblock {\em TrAC - Trends in Analytical Chemistry}, 21(9-10):698--708, 2002.

\bibitem{Thomas}
C.~L.~P. Thomas.
\newblock {D3.1 Prototype sampling system for reproducible non-invasive
  clinical sampling protocol.}
\newblock 2016.

\bibitem{Salek2013}
Reza~M. Salek, Christoph Steinbeck, Mark~R. Viant, Royston Goodacre, and
  Warwick~B. Dunn.
\newblock {The role of reporting standards for metabolite annotation and
  identification in metabolomic studies}.
\newblock {\em GigaScience}, 2(1), 2013.

\bibitem{Shrivastava2016}
Abhinav Shrivastava, Abhinav Gupta, and Ross Girshick.
\newblock {Training region-based object detectors with online hard example
  mining}.
\newblock In {\em Proceedings of the IEEE Computer Society Conference on
  Computer Vision and Pattern Recognition}, volume 2016-December, pages
  761--769, 2016.

\bibitem{Liu2016}
Wei Liu, Dragomir Anguelov, Dumitru Erhan, Christian Szegedy, Scott Reed,
  Cheng~Yang Fu, and Alexander~C. Berg.
\newblock {SSD: Single shot multibox detector}.
\newblock In {\em Lecture Notes in Computer Science (including subseries
  Lecture Notes in Artificial Intelligence and Lecture Notes in
  Bioinformatics)}, volume 9905 LNCS, pages 21--37, 2016.

\end{thebibliography}

\end{document}